\documentclass{article}

\usepackage{microtype}
\usepackage{graphicx}
\usepackage{booktabs} 
\usepackage[inline]{enumitem}
\usepackage[]{subfig}
\usepackage{hyperref}
\usepackage{multirow}
\usepackage{graphicx}

\usepackage[accepted]{icml2025}

\usepackage{amsmath}
\usepackage{amssymb}
\usepackage{mathtools}
\usepackage{amsthm}

\usepackage[capitalize,noabbrev]{cleveref}

\theoremstyle{plain}
\newtheorem{theorem}{Theorem}[section]

\newtheorem{lemma}[theorem]{Lemma}

\theoremstyle{definition}

\theoremstyle{remark}

\usepackage[textsize=tiny]{todonotes}

\usepackage{xspace}
\usepackage{pifont}
\usepackage{dblfloatfix}
\newcommand{\methodAbbr}{D$^3$HR\xspace}
\newcommand{\red}[1]{{\color{red}#1}}

\usepackage{color, colortbl}
\definecolor{LightCyan}{rgb}{0.88,1,1}

\newcommand{\onedot}{\ifx\@let@token.\else.\null\fi\xspace}

\newcommand{\eg}[0]{\emph{e.g}\onedot}

\icmltitlerunning{Taming Diffusion for Dataset Distillation with High Representativeness}

\begin{document}

\twocolumn[
\icmltitle{Taming Diffusion for Dataset Distillation with High Representativeness}




\begin{icmlauthorlist}
\icmlauthor{Lin Zhao}{yyy}
\icmlauthor{Yushu Wu}{yyy}
\icmlauthor{Xinru Jiang}{yyy}
\icmlauthor{Jianyang Gu}{sch}
\icmlauthor{Yanzhi Wang}{yyy}
\icmlauthor{Xiaolin Xu}{yyy}
\icmlauthor{Pu Zhao}{yyy}
\icmlauthor{Xue Lin}{yyy}
\end{icmlauthorlist}

\icmlaffiliation{yyy}{Northeastern University}
\icmlaffiliation{sch}{The Ohio State University}
\icmlcorrespondingauthor{Pu Zhao}{p.zhao@northeastern.edu}
\icmlcorrespondingauthor{Xue Lin}{xue.lin@northeastern.edu}

\icmlkeywords{Machine Learning, ICML}

\vskip 0.3in
]



\printAffiliationsAndNotice{}  

\begin{abstract}
Recent deep learning models demand larger datasets, driving the need for dataset distillation to create compact, cost-efficient datasets while maintaining performance.
Due to the powerful image generation capability of diffusion, it has been introduced to this field for generating distilled images.
In this paper, we systematically investigate issues present in current diffusion-based dataset distillation methods, including inaccurate distribution matching, distribution deviation with random noise, and separate sampling.
Building on this, we propose \methodAbbr, a novel diffusion-based framework to generate distilled datasets with high representativeness.
Specifically, we adopt DDIM inversion to map the latents of the full dataset from a low-normality latent domain to a high-normality Gaussian domain, preserving information and ensuring structural consistency to generate representative latents for the distilled dataset.
Furthermore, we propose an efficient sampling scheme to better align the representative latents with the high-normality Gaussian distribution.
%
Our comprehensive experiments demonstrate that \methodAbbr can achieve higher accuracy across different model architectures compared with state-of-the-art baselines in dataset distillation. Source code: \url{https://github.com/lin-zhao-resoLve/D3HR}.
\end{abstract}

\section{Introduction}


%
Driven by the scaling law, recent deep learning models have expanded in scale, demanding exponentially larger datasets for optimal training performance, which makes dataset maintenance costly and labor-intensive~\cite{deng2009imagenet,alexey2020image}.
%
To achieve both computational and storage efficiency, dataset distillation generates a small distilled dataset to replace the full dataset for training~\cite{wang2018dataset,loolarge,lee2024selmatch}, while targeting comparable performance as the full dataset.
In contrast to dataset pruning methods (selecting a subset from the full dataset)~\cite{zhang2024spanning,sorscher2022beyond,tan2024data}, dataset distillation treats the distilled dataset as continuous parameters that are optimized to integrate 
information from the full dataset.
This approach achieves better performance, especially with relatively high compression rates.~\cite{,du2024diversity,sun2024diversity,yin2023squeeze}.



Recent data distillation methods primarily focus on improving the efficiency of the distillation algorithm~\cite{yin2023squeeze,sun2024diversity,Su_2024_CVPR,gu2024efficient}.
Among them, the diffusion-based methods achieve strong performance due to their remarkable generative capabilities.
%
%
They capture the most informative features of the full dataset by extracting representative latents from the pre-trained Variational Autoencoder (VAE) latent space \cite{Su_2024_CVPR,gu2024efficient}.
Thus, the architectural dependency during   distillation   is eliminated, enabling a one-time generation cost for training various model architectures.
This strategy enhances the efficiency of distillation and significantly improves cross-architecture generalization.

\begin{figure}[]
    \captionsetup[subfloat]{captionskip=-6pt}
    \centering
    \vspace{-2em}
    \hfill
    \subfloat[][Low-normality (VAE)]
    {\includegraphics[width=0.43\linewidth]{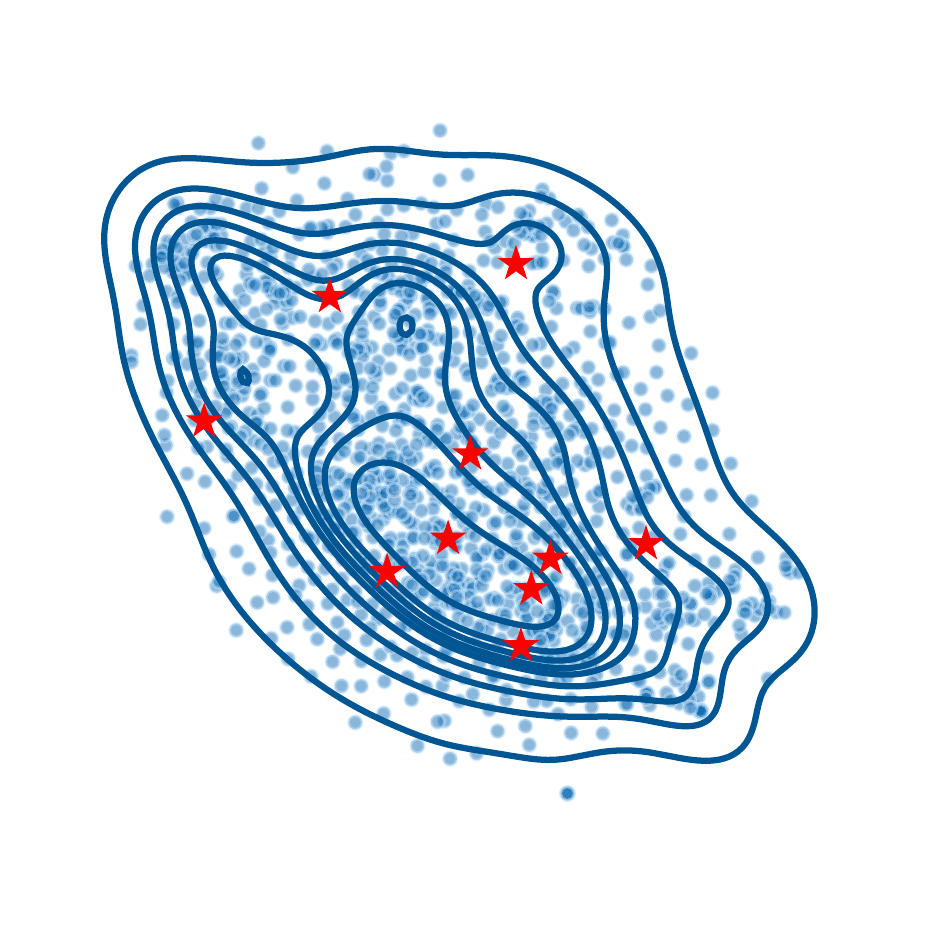}\label{fig:vae-space-normality}}
    \hfill
    \subfloat[][High-normality(mapped)]
    {\includegraphics[width=0.43\linewidth]{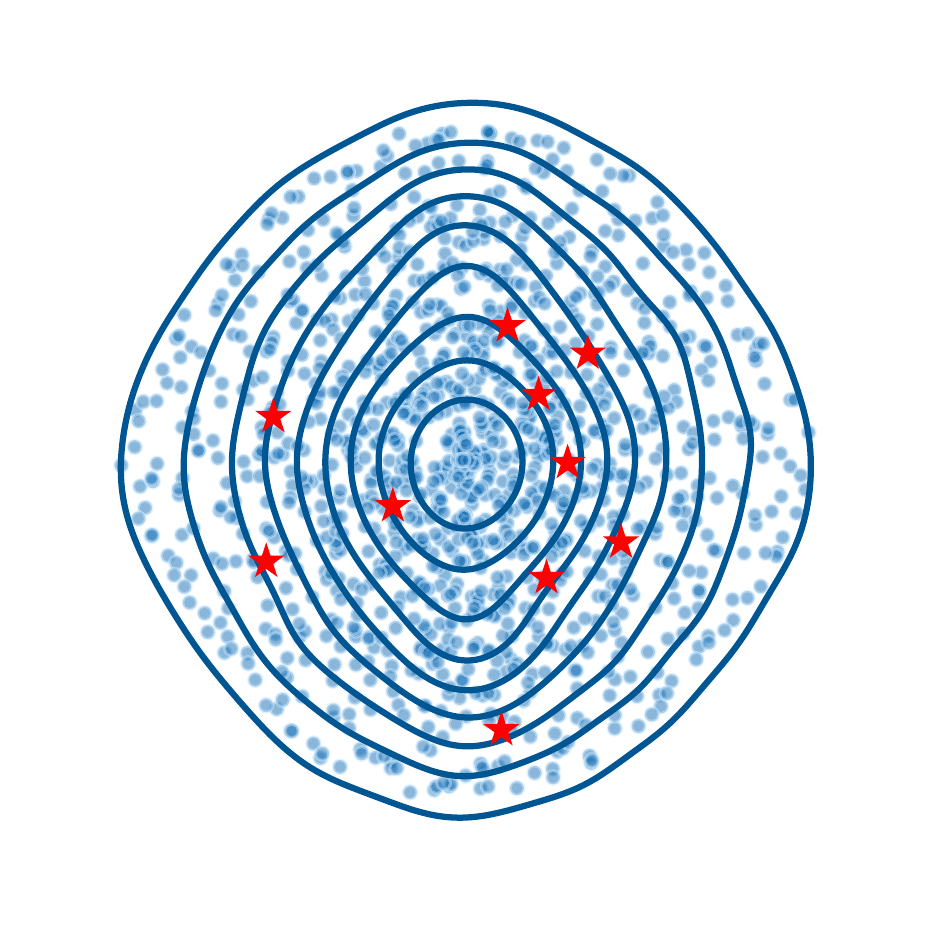}\label{fig:noise-space-normality}}
    \hfill
    \caption{\textbf{t-SNE visualization of the low-normality VAE space and high-normality noise space for class “Goldfish".} The blue contour lines are the probability density curves of the distribution using kernel density estimation, highlighting the structure and concentration of the latents (blue dots).
    \red{\ding{72}} in (b) marks the $10$ representative latents generated in the noise space, corresponding to \red{\ding{72}} in (a) after DDIM sampling, which preserves the structure of VAE space and concentrates in high-density regions.}
    \vspace{-4mm}
    \label{fig:space-normality}
\end{figure}

However, the above diffusion-based methods may struggle to guarantee the representativeness of the distilled dataset.
\begin{enumerate*}[label=(\roman*)]
    \item
    \textbf{Inaccurate distribution matching.}
    This challenge arises from leveraging diffusion models to generate distilled datasets, relying on accurately modeling the VAE feature distributions for proper decoding.
    However, as shown in \cref{fig:vae-space-normality}, the low-degree of normality\footnote{Normality refers to the degree of the latent space data conforms to a normal distribution. A higher degree of normality indicates closer conformity.}  for the distribution in VAE latent space results in difficulties in effectively matching  the distribution.
    \item 
    \textbf{Distribution deviation with random noise.} 
    Besides, these methods generate distilled images from the initial noise, which inject unpredictable randomness to latents, potentially violating the structural and representative information gathered in the VAE space with potential distribution deviation.
    \item \textbf{Separate sampling.} 
    Moreover, the distilled data are individually matched to parts of the entire VAE distribution without considering the overall distribution of the distilled dataset, which may lead to an incomplete representation of the desired distribution.
\end{enumerate*}

Inspired by these insights, 
we propose a novel framework, Taming \textbf{D}iffusion for \textbf{D}ataset \textbf{D}istillation with \textbf{H}igh \textbf{R}epresentativeness (\textbf{D$^3$HR}).
We reveal an optimal strategy for modeling the VAE latent domain  by mapping it to a simpler distribution domain while preserving fundamental information and ensuring structural consistency.
Toward this end, we adopt the deterministic DDIM inversion to map latents from the VAE space to the noise space with higher 
normality, so that a Gaussian distribution can accurately match the noised latent distribution, as illustrated in \cref{fig:noise-space-normality}.
Then we sample a set of latents from the estimiated Gaussian distribution and convert the latents back to VAE space by DDIM sampling for generating the distilled dataset.
To ensure the distribution matching between the distilled dataset and the full dataset, we incorporate a new constraint with our group sampling method, which tights the similarity between the distribution of the generated latents and that of the original latents. 

Our contributions can be summarized as follows:
\begin{itemize}[topsep=1pt, itemsep=1pt, parsep=1pt]
    \item We propose \methodAbbr, a dataset distillation framework leveraging DDIM inversion to map the original data distribution to a Gaussian distribution, 
    which makes it easier for distribution matching (better normality) with less randomness (bijective DDIM mapping to preserve structural information), 
    effectively addressing  inaccurate distribution matching and distribution deviation  in 
     previous diffusion-based methods.
    \item We propose a novel sampling scheme to efficiently align the distributions of generated distilled latents and  original latents from the full dataset, overcoming the shortcoming of individual sampling. 
    \item We conduct extensive experiments and  demonstrate that  \methodAbbr significantly outperforms state-of-the-art dataset distillation methods, with detailed ablation studies and analysis. 
\end{itemize}

\section{Related Work}

\noindent \textbf{Dataset Distillation}.~
Dataset distillation \cite{wang2018dataset} 
is designed to improve training efficiency by generating a small dataset to replace the original large dataset for training.
Previous methods \cite{zhao2020dataset,cazenavette2022dataset}  employ a bi-level optimization process, which updates the model parameters and the distilled dataset simultaneously.
According to the different matching strategies  in the distillation, the methods can be classified into:
gradient matching \cite{zhao2020dataset,zhao2021dataset,lee2022dataset}, distribution matching \cite{wang2022cafe,zhao2023dataset,zhao2023improved,zhang2023echo,deng2024exploiting}, and trajectory matching \cite{cazenavette2022dataset,cui2023scaling,du2023minimizing,du2024sequential}.

Recent methods demonstrate that the bilevel optimization approach is time-consuming and does not scale well for large datasets \cite{loolarge,sun2024diversity,yin2023squeeze,shao2024generalized,yu2024teddy}. 
To address this,  more efficient dataset distillation methods are developed. 
%
SRe$^2$L \cite{yin2023squeeze} and DWA \cite{du2024diversity} use the BN layers of the pre-trained teacher model as supervisory information and synthesize each distilled image individually to accelerate the generation.
%
%
RDED \cite{sun2024diversity} emphasizes the importance of image realism and generates images by stitching patches selected by the teacher model.
However, these methods rely on BN layers or the data distribution of other layers from the teacher, which limits their applicability to different model architectures.
%
%

Unlike the above methods, the diffusion-based methods leverage the powerful generative capabilities of diffusion models, eliminating the dependency on  teacher models.
Specifically, D\textsuperscript{4}M~\cite{Su_2024_CVPR} 
generates the distilled dataset by clustering the latents of the full dataset in the VAE latent space. 
\citet{gu2024efficient} reconstruct the training loss of diffusion model in the VAE space, aiming to generate images representing the full dataset. 
However, it requires fine-tuning multiple models for large datasets.
The above methods struggle to accurately capture representative latents that match the full dataset distribution. 

\noindent \textbf{Diffusion Models}.~
Diffusion models target approximating the data distribution $q_\theta(x_\theta)$ with a learned model distribution $p_\theta(x_\theta)$.
Denoising diffusion probabilistic models~(DDPMs)~\cite{ho2020denoising} optimizes a variational lower bound based on the Markov Chain, allowing models to generate high-quality samples. 
DDIM~\cite{song2020denoising} built on  DDPMs provides a more efficient and deterministic sampling method by removing the randomness in each reverse step, thus offering a faster and more controlled generation.  
Due to the deterministic approach, DDIM sampling can be inverted, allowing users to map their generated or real image back to its corresponding noise representation.
Previous works~\cite{rombach2022high} rely on the UNet architecture, 
and recent works~\cite{peebles2023scalable,esser2024scaling,shen2024lazydit} demonstrate the superior performance of Diffusion Transformer~(DiT) in image generation by introducing Transformer architecture to diffusion models.
\section{Background, Formulation, and Motivation}
%

\subsection{Preliminaries on Diffusion Models} \label{sec:background}
\textit{Diffusion Models}~\cite{ho2020denoising} are proposed to generate high-quality images by transforming the random Gaussian noise $x_t$ into the image $x_0$ through a discrete Markov chain.
Among them, Latent Diffusion Model~\cite{rombach2022high} enhances efficiency by leveraging VAE to compress the pixel space $\mathcal{X}_0$ into the latent space $\mathcal{Z}_0: z_0 = E(x_0)$, and decoding the latents back to images at the end of diffusion backward process: $x_0 = D(z_0)$.
During training, the forward process adds random noise to the initial latent $z_0$:
\begin{equation}
    z_t = \sqrt{\alpha_t} z_0 + \sqrt{1 - \alpha_t} \epsilon, \quad \text{with} \quad \epsilon \sim \mathcal{N}(0, \mathbf{I}),
\label{equ:ddpm}
\end{equation}
where $\alpha_t$ is a hyper-parameter and $z_t$ represents the noise at timestep $t$.
%
%
During inference, the backward process iteratively removes  noise in $z_t$ through  schedulers to get $z_0$.

\textit{Denoising Diffusion Implicit Models}~(DDIM) \cite{song2020denoising} 
introduces a deterministic sampling method by configuring the variance of the distribution at each step, enabling a one-to-one mapping from $z_t$ to $z_0$, as expressed below:
\begin{equation}
\scalebox{1.0}{$
\begin{aligned}
    z_{t-1} = & \sqrt{\frac{\alpha_{t-1}}{\alpha_{t}}} z_t + \\
     & \sqrt{\alpha_{t-1}} \left( \sqrt{\frac{1}{\alpha_{t-1}}-1} - \sqrt{\frac{1}{\alpha_{t}}-1} \right) \varepsilon_{\theta} (z_t, t, \mathcal{C}).
\end{aligned}$}
\label{equ:ddim}
\end{equation}
where $\varepsilon_{\theta} (z_t, t, \mathcal{C})$  is a function with trainable parameters $\theta$, $\mathcal{C}$ denotes the class condition.


\subsection{Problem Formulation for Dataset Distillation}
Dataset distillation aims to synthesis a small distilled dataset $\mathcal{S} = \left\{ \mathbf{\hat{x}}_i, \hat{y}_i \right\}_{i=1}^{N_\mathcal{S}}$ to replace the full dataset $\mathcal{F} = \left\{ \mathbf{x}_i, y_i \right\}_{i=1}^{N_\mathcal{F}}$ for training, where $N_\mathcal{S} \ll N_\mathcal{F}$,  $\mathbf{\hat{x}}_i$ \& $\mathbf{x}_i$ are images, and $\hat{y}_i$ \& $y_i$ are labels.
To maintain high performance with a smaller $\mathcal{S}$, Algorithm $\mathcal{A}$ is proposed to address the problem of generating $\mathcal{S}$ from $\mathcal{F}$: $\mathcal{S} = \mathcal{A}(\mathcal{F}) \mid N_\mathcal{S} \ll N_\mathcal{F}$.
It is expected that training a model on $\mathcal{S}$ achieves a comparable performance to training on $\mathcal{F}$,  assuming that $\mathcal{S}$ encapsulates substantial information from the original $\mathcal{F}$.



\subsection{Motivation}
\label{sec:motivation}

Existing dataset distillation works have certain limitations as discussed below, which motivate us to develop our \methodAbbr. 
 
\textbf{Scalability to multiple model architectures}.~ 
In some dataset distillation works \cite{yin2023squeeze,loolarge}, a teacher model $T$ trained by $\mathcal{F}$ is used to guide the distillation process: $\mathcal{S} = \mathcal{A}(\mathcal{F}, T)$.
It relies on a specific model architecture.
A new model architecture requires distilling another dataset with the new teacher architecture expensively trained on the full dataset, making it hard to scale.  

To address this problem, recent works \cite{gu2024efficient,Su_2024_CVPR} leverage the powerful generative capabilities of diffusion models $G$ \cite{azizi2023synthetic} to produce $\mathcal{S}$ without the guidance from a specific teacher model $T$: $\mathcal{S} = \mathcal{A}(\mathcal{F}, G)$.
In this setup, once $\mathcal{S}$ is generated with just a \textbf{one-time cost}, it can be applied to various model architectures (\eg, ResNet, EfficientNet, and VGG), instead of generating multiple $\mathcal{S}$ for each model architecture.

\textbf{Inaccurate distribution matching in the latent space}.
There are still some limitations for the current diffusion-based data distillation methods.
Specifically, the original images of the class $\mathcal{C}$ are converted to latents $\mathcal{Z}_{0,\mathcal{C}} = \left\{ z_0^i \mid \mathcal{C} \right\}_{i=1}^{N_{\mathcal{F},\mathcal{C}}}$ through the VAE, and the methods propose to find and describe the data distributions in the latent space, 
for further  synthesis of $n$ latents for  $\mathcal{C}$ (i.e., $n$ images-per-class (IPC)), hoping that $n$ latents can   match the real data distributions in the latent space. For example, 
\citet{Su_2024_CVPR} apply K-means to cluster $\mathcal{Z}_{0,\mathcal{C}}$ into $n$ groups to get the centroids for better synthesis of $n$ latents.
\citet{gu2024efficient} minimize the cosine similarity between each synthesized latent and a random subset $\mathcal{Z}_s$, where \( \mathcal{Z}_s \subset \mathcal{Z}_{0,\mathcal{C}} \).

However, the distribution of the original dataset in the latent space is too complex to accurately describe or match. It is a multi-component Gaussian mixture distribution as below. 
\begin{lemma}
Each latent in VAE latent space is randomly sampled from a distinct component of a multi-component Gaussian mixture distribution.
\end{lemma}
The proof is shown in \cref{sec:Theoretical}.1.  $\mathcal{Z}_{0,\mathcal{C}}$ is a multi-component Gaussian mixture distribution and hard-to-fit, with examples shown in \cref{fig:vae-space-normality}. 

Therefore, it is challenging to generate representative latents in the complex $\mathcal{Z}_{0,\mathcal{C}}$. 
The K-means algorithm \cite{Su_2024_CVPR} assumes that each cluster is spherical for easier clustering $\mathcal{Z}_{0,\mathcal{C}}$,  which does not align with the more complex distribution of $\mathcal{Z}_{0,\mathcal{C}}$.
The cosine similarity    \cite{gu2024efficient}   focuses on the direction to a subset of the  distribution, and ignores the distance, making it less effective to capture the differences in probability density within $\mathcal{Z}_{0,\mathcal{C}}$.

\begin{figure*}[t]
    \centering
    \includegraphics[width=1.\linewidth]{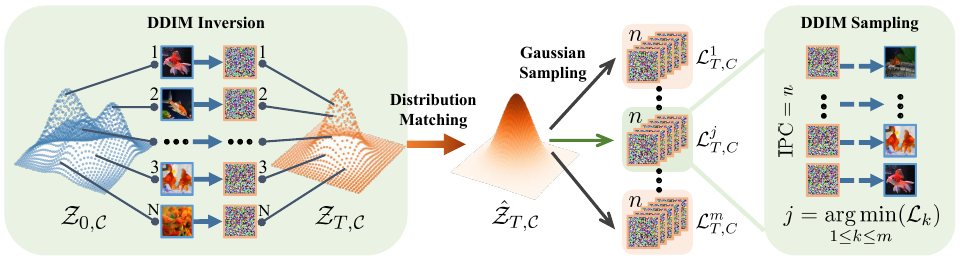}
    \caption{
    \textbf{Illustration of proposed \methodAbbr framework.}
    With the latents from the VAE, DDIM inversion is applied to map the latent embeddings to a  Gaussian domain with better normality, which can further be matched to a Gaussian distribution.
    Then, we follow~\cref{subsec:gaussian sampling} to sample representative latents based on different IPC requirements, and generate true images through DDIM sampling.
    }
    \label{fig:framework}
\end{figure*}

\textbf{Distribution deviation with random noise}.~
Current methods generate images from the initial noise.
They obtain the noise either by adding random noise to the representative latents in the VAE space via \cref{equ:ddpm}  \cite{Su_2024_CVPR}, or by generating it randomly, with the representative latents obtained by multiple fine-tuned diffusion models \cite{gu2024efficient}. 
However, due to  the inherent randomness of   initial noise, the distribution of denoised latents may deviate or shift from that of the sampled latents before adding noise, as shown in Appendix B-\cref{fig:d4m-shift}, leading to potential accuracy loss after training. 

\textbf{Separate sampling}.~  
Additionally, current diffusion-based works generate representative latents without considering their overall distribution. Due to the limited number of samples, although each latent may be matched to parts of the desired distribution $\mathcal{Z}_{0,\mathcal{C}}$, the overall distribution of the sampled $n$ latents may not closely align with $\mathcal{Z}_{0,\mathcal{C}}$, resulting in an under-represented distilled dataset. 


\section{Methodology for Dataset Distillation}
Motivated by the limitations, we present \methodAbbr, a diffusion-based dataset distillation framework via domain mapping. 
%
%

\subsection{Framework Overview}\label{subsec:overview}

Our framework has three stages: domain mapping, distribution matching, and group sampling. 
As shown in \cref{fig:framework}, we first convert images to latents through VAE and map the VAE latent domain to the noise domain via DDIM inversion.
Next, with distribution matching, we use a Gaussian distribution to match the mapped noised latent distribution   from the previous stage. 
Finally, in the group sampling stage, we sample a number of latents from the Gaussian distribution and convert the latents back to images to generate the distilled dataset.  
The algorithm is shown in Algorithm~\ref{algo:method} with details introduced in the following subsections.

\begin{algorithm}[tb]
\small
   \caption{\methodAbbr Algorithm}
   \label{alg:base method}
\begin{algorithmic}
   \STATE {\bfseries Input:} Full dataset $\mathcal{F}=(X,Y)$, VAE encoder-decoder $E, D$, Inverse steps $T$, IPC number ${n}$
   \STATE {\bfseries Output:} Distilled dataset $\mathcal{S}$
   \STATE Initialize $\mathcal{S} = \emptyset$
   \FOR{\textbf{each} $class\ \mathcal{C}$}
   \STATE $y_{\mathcal{C}} \gets$ $label\ of\ class\ \mathcal{C}$
   \STATE $\mathcal{Z}_{0,\mathcal{C}}=E(X|C)$
   \STATE Initialize $\mathcal{Z}_{T,\mathcal{C}} = \emptyset$ 
   \FOR{\textbf{each} $z_0\in \mathcal{Z}_{0,\mathcal{C}}$}
       \STATE Update $z_0$ to $z_T$ via \cref{equ:ddim_inversion} with $T$ steps
       \STATE \quad \quad \quad \quad \quad \quad \quad \quad \quad \quad \quad \quad \quad \text{\(\triangleright\)} \text{DDIM Inversion}
       \STATE $\mathcal{Z}_{T,\mathcal{C}} \gets \mathcal{Z}_{T,\mathcal{C}} \cup \{z_T\}$
   \ENDFOR
   \STATE $\mu_{T,\mathcal{C}} \gets$ $mean\ of$ $\mathcal{Z}_{T,\mathcal{C}}$
   \STATE $\sigma^2_{T,\mathcal{C}} \gets$ $variance\ of$ $\mathcal{Z}_{T,\mathcal{C}}$
   \STATE $\hat{\mathcal{Z}}_{T,\mathcal{C}} \gets \mathcal{N}(\mu_{T,\mathcal{C}}, \sigma^2_{T,\mathcal{C}})$
   \STATE 
   Initialize $\mathcal{L}_{T,\mathcal{C}} = \emptyset$ \, \, \, \quad \quad \quad \quad  \quad \text{\(\triangleright\)} \text{Group Sampling}
   \FOR{$k \gets 1$ to ${m}$}
       \STATE Gaussian Sample ${n}$   latents to get $\mathcal{R}^k = \{\hat{z}^i_{T,\mathcal{C}}\}_{i=1}^{n}$ 
       \STATE Calculate $\mathcal{L}^k_{T,\mathcal{C}}$ via \cref{equ:L} for $\mathcal{R}^k$
       \STATE $\mathcal{L}_{T,\mathcal{C}} = \mathcal{L}_{T,\mathcal{C}} \cup \{\mathcal{L}^k_{T,\mathcal{C}}\}$
   \ENDFOR
   \STATE $j = \underset{1 \leq k \leq {m}}{\arg\min}(\mathcal{L}^k_{T,\mathcal{C}})$
   \FOR{\textbf{each} $\hat{z}_T\in \mathcal{R}^j$}
       \STATE Update $\hat{z}_T$ to $\hat{z}_0$ via \cref{equ:ddim} in $T$ steps 
       \STATE \quad \quad \quad \quad \quad \quad \quad \quad \quad \quad \quad \quad \quad  \text{\(\triangleright\)} \text{DDIM Sampling}
       \STATE $\mathcal{S} \gets \mathcal{S} \cup \{D(\hat{z}_0), y_{\mathcal{C}}\}$ 
   \ENDFOR
   \ENDFOR
   \STATE {\bfseries return} $\mathcal{S}$ 
\label{algo:method}
\end{algorithmic}
\end{algorithm}

\subsection{Domain Mapping}\label{subsec:domain mapping}
It is essential that the small distilled dataset can sufficiently represent the original large data.
Previous works \cite{gu2024efficient,Su_2024_CVPR} use the distribution in VAE latent space $\mathcal{Z}_{0,\mathcal{C}} = \left\{ z_0^i \mid \mathcal{C} \right\}_{i=1}^{N}$  to distill the  information of class $\mathcal{C}$ from the original dataset. 
As discussed in \cref{sec:motivation}, it is challenging to generate \({n}\) latents to match the VAE latent space $\mathcal{Z}_{0,\mathcal{C}}$. 
To  make it easier for distribution matching, 
we further map the distribution in  $\mathcal{Z}_{0,\mathcal{C}}$ to a high-normality Gaussian space $\mathcal{Z}_{T,\mathcal{C}}$ by \textit{DDIM inversion}  \cite{dhariwal2021diffusion} for each class  $\mathcal{C}$, as illustrated in \cref{fig:framework}.
Specifically, for each latent $z_{0}$ in class $\mathcal{C}$,  we perform a few DDIM inversion steps as follows,
\begin{equation}
\scalebox{1.0}{$
\begin{aligned}
    z_{t+1} = & \sqrt{\frac{\alpha_{t+1}}{\alpha_{t}}} z_t + \\
       &{\alpha_{t+1}} \left( \sqrt{\frac{1}{\alpha_{t+1}}-1} - \sqrt{\frac{1}{\alpha_{t}}-1} \right) \varepsilon_{\theta} (z_t, t, \mathcal{C}).
\end{aligned}$}
\label{equ:ddim_inversion}
\end{equation}

\textit{DDIM inversion} is based on the assumption that the ODE process of DDIM sampling can be inverted in a few steps.
There are other choices for domain mapping such as directly adding random Gaussian noise through the forward of Denoising Diffusion Probabilistic Models (DDPM)  in \cref{equ:ddpm}.
However, DDPM inherently introduces randomness in the mapping, leading to difficulties for accurately describing the distribution with potential distribution shift or information loss, as presented in \cref{fig:ddpm-space-normality}. 
Different from DDPM, DDIM inversion with \cref{equ:ddim_inversion} offers the key prerequisites for obtaining a representative subset in $\mathcal{Z}_{T,\mathcal{C}}$: 
(i) \textbf{Information Preservation:} The mapping between two domains is bijective because of the deterministic process, where each element in $\mathcal{Z}_{T,\mathcal{C}}$ directly corresponds to an element in $\mathcal{Z}_{0,\mathcal{C}}$ and vice versa, which avoids the loss of key features during the mapping process.
(ii) \textbf{Structural Consistency:} The latents in $\mathcal{Z}_{T,\mathcal{C}}$ can retain the structural information of $\mathcal{Z}_{0,\mathcal{C}}$, 
ensuring the structural consistency and distribution alignment.
%


%
%
%

\subsection{Distribution Matching}\label{subsec:distribution_matching}

With DDIM inversion, we  can obtain   a   discrete distribution $\mathcal{Z}_{T,\mathcal{C}} = \left\{ z_T^i \mid \mathcal{C} \right\}_{i=1}^{N} $.  We have the following lemma. 
\begin{lemma}
\label{lemma:ddim}
For DDIM inversion with $T$ steps, with sufficiently large  $T$, $\mathcal{Z}_{T,\mathcal{C}}$ can be approximated as a Gaussian distribution.
\end{lemma}
The proof is shown in \cref{sec:Theoretical}.2. 
Since the latents in 
$\mathcal{Z}_{T,\mathcal{C}}$ can be interpreted as independently and identically Gaussian distributed (i.i.d.) latents,  we approximate     $\mathcal{Z}_{T,\mathcal{C}}$ with a   Gaussian distribution $\mathcal{\hat{Z}}_{T,\mathcal{C}}$ based on the law of large numbers~\cite{hsu1947complete}. 
%
To obtain the statistical properties of $\mathcal{\hat{Z}}_{T,\mathcal{C}}$, we compute the the mean $\mu_{T,\mathcal{C}}$ and variance $\sigma^2_{T,\mathcal{C}}$  from  $\mathcal{Z}_{T,\mathcal{C}}$ as  the mean and variance of  $\mathcal{\hat{Z}}_{T,\mathcal{C}}$. 

Thus, we can obtain the Gaussian distribution 
$\mathcal{\hat{Z}}_{T,\mathcal{C}}\sim\mathcal{N}(\mu_{T,\mathcal{C}}, \sigma^2_{T,\mathcal{C}})$ 
 in the latent space to represent the class $\mathcal{C}$. Since the dimensions in the noise space of DDIM inversion are independent \cite{song2020denoising}, the probability density function (PDF) of  $\mathcal{\hat{Z}}_{T,\mathcal{C}}$ can be expressed as the following,
\begin{equation}
\scalebox{0.9}{$
\begin{split}
    f(\mathbf{\hat{z}}_{T,\mathcal{C}}) = \prod_{i=1}^{d} \frac{1}{\sqrt{2\pi (\sigma_{T,\mathcal{C}}^i)^2}} \exp\left(-\frac{(\hat{z}_{T,\mathcal{C}}^i - \mu_{T,\mathcal{C}}^i)^2}{2(\sigma_{T,\mathcal{C}}^i)^2}\right), 
\end{split}
$}
\label{equ:kd}
\end{equation}
where $\mu_{T,\mathcal{C}}^i$ and $\sigma_{T,\mathcal{C}}^i$ denote the mean and standard deviation for each dimension of $\mu_{T,\mathcal{C}}$  and   $\sigma_{T,\mathcal{C}}$, respectively.   

%

%

\begin{table*}[ht]
\centering
\resizebox{\textwidth}{!}{
\begin{tabular}{llcccccccccc}
\toprule
\multirow{2}{*}{Dataset}& \multirow{2}{*}{IPC} & \multicolumn{5}{c}{ResNet-18}
& \multicolumn{5}{c}{ResNet-101}  \\
\cmidrule(lr){3-7} \cmidrule(lr){8-12} 
 & & SRe$^2$L & DWA & D$^4$M & RDED & Ours 
 & SRe$^2$L & DWA & D$^4$M & RDED & Ours\\
\midrule
\multirow{2}{*}{CIFAR-10} & 10 & $29.3 \pm 0.5$ & $32.6\pm0.4$ & $33.5$ & $37.1 \pm 0.3$ & $\textbf{41.3} \pm \textbf{0.1}$ 
& $24.3 \pm 0.6$ & $25.2 \pm 0.2$ & $29.4$ & $33.7 \pm 0.3$ & $\textbf{35.8} \pm \textbf{0.6}$\\
 & 50 & $45.0 \pm 0.7$ & $53.1\pm0.3$ & $62.9$ & $62.1 \pm 0.1$ & $\textbf{70.8} \pm \textbf{0.5}$ 
 & $34.9 \pm 0.1$ & $48.2 \pm 0.4$ & $54.4$ & $51.6 \pm 0.4$ & $\textbf{63.9} \pm \textbf{0.4}$\\
\midrule
\multirow{2}{*}{CIFAR-100} & 10 & $31.6 \pm 0.5$ & $39.6\pm0.6$ & $38.1$ & $42.6 \pm 0.2$ & $\textbf{49.4} \pm \textbf{0.2}$ 
& $30.7\pm0.3$ & $35.9\pm0.5$ & $33.3$ & $41.1\pm0.2$ & $\textbf{46.0} \pm \textbf{0.5}$\\
 & 50 & $52.2 \pm 0.3$ & $60.9\pm0.5$ & $63.2$ & $62.6 \pm 0.1$ & $\textbf{65.7} \pm \textbf{0.3}$ 
 & $56.9\pm0.1$ & $58.9\pm0.6$ & $64.7$ & $63.4\pm0.3$ & $\textbf{66.6} \pm \textbf{0.2}$\\
\midrule
\multirow{3}{*}{Tiny-ImageNet} & 10 & $16.1 \pm 0.2$ & $40.1 \pm 0.3$ & $34.1$ & $41.9 \pm 0.2$ & $\textbf{44.4} \pm \textbf{0.1}$ 
& $14.6\pm1.1$ & $38.5\pm0.7$ & $33.4$ & $22.9\pm3.3$ & $\textbf{43.2} \pm \textbf{0.5}$\\
& 50 & $41.4 \pm 0.4$ & $52.8\pm0.2$ & $46.2$ & $ \textbf{58.2}\pm \textbf{0.1}$ & $56.9\pm0.2$ 
& $42.5\pm0.2$ & $54.7\pm0.3$ & $51.0$ & $41.2\pm0.4$ & $\textbf{59.4} \pm \textbf{0.1}$\\
 & 100 & $49.7 \pm 0.3$ & $56.0\pm0.2$ & $51.4$ & - & $\textbf{59.3} \pm \textbf{0.1}$ 
 & $51.5\pm0.3$ & $57.4\pm0.3$  & $55.3$ & -& $\textbf{61.4} \pm \textbf{0.1}$\\
\midrule
\multirow{3}{*}{ImageNet-1K}
 & 10 & $21.3 \pm 0.6$ & $37.9\pm0.2$ & $27.9$ & $42.0 \pm 0.1$ & $\textbf{44.3} \pm \textbf{0.3}$ 
 & $30.9\pm0.1$ & $46.9\pm0.4$ & $34.2$ & $48.3\pm1.0$ & $\textbf{52.1} \pm \textbf{0.4}$\\
 & 50 & $46.8 \pm 0.2$ & $55.2\pm0.2$ & $55.2$ & $56.5 \pm 0.1$ & $\textbf{59.4} \pm \textbf{0.1}$ 
 & $60.8\pm0.5$ & $63.3\pm0.7$ & $63.4$ & $61.2\pm0.4$ & $\textbf{66.1} \pm \textbf{0.1}$\\
 & 100 & $52.8 \pm 0.3$ & $59.2\pm0.3$ & $59.3$ & - & $\textbf{62.5}\pm \textbf{0.0}$ 
 & $62.8\pm0.2$ & $66.7\pm0.2$ & $66.5$ & - & $\textbf{68.1} \pm \textbf{0.0}$\\
\bottomrule
\end{tabular}
}
\caption{\textbf{Comparison of Top-1 accuracy across four datasets with other methods.} Following SRe$^2$L \cite{yin2023squeeze}, all methods use ResNet-18 as teacher model, and train separately on ResNet-18 and ResNet-101 with the soft label. As D$^4$M \cite{Su_2024_CVPR} and our \methodAbbr 
do not need teacher models, we just  use the soft label as supervision. 
For $\text{IPC}=100$, some results are unavailable with `-' due to  their inappropriate default parameter settings. 
}
\label{tab:performance_comparison}
\end{table*}

\subsection{Group Sampling}\label{subsec:gaussian sampling}
%

With the Gaussian distribution $\mathcal{\hat{Z}}_{T,\mathcal{C}}$ for the class $\mathcal{C}$, next we generate the distilled dataset. 
Specifically, we generate   $n$ representative latents for each class $\mathcal{C}$ by capturing the statistic characteristics of $\mathcal{\hat{Z}}_{T,\mathcal{C}}$.
Using the Ziggurat algorithm~\cite{marsaglia2000ziggurat}, we can randomly sample $n$  i.i.d. latents probabilistically following $\mathcal{\hat{Z}}_{T,\mathcal{C}}$ with $f(\mathbf{\hat{z}}_{T,\mathcal{C}})$, which together form a   subset.  However, although each latent is sampled from $\mathcal{\hat{Z}}_{T,\mathcal{C}}$, due to the limited number of samples ($n$) for distillation, the overall distribution of the subset with $n$ samples may still deviate from the desired $\mathcal{\hat{Z}}_{T,\mathcal{C}}$, leading to certain performance degradation. 
 


To address this problem and further improve the quality of the distilled dataset, we propose to sample numerous random subsets (each subset consists of $n$ i.i.d. latents) and efficiently search for the most representative subset as the final subset.  Specifically, by repeating the above sampling with the Ziggurat algorithm multiple times, $m$ random subsets are available with varying statistical distributions.
To search for the most representative subset, we further propose an efficient algorithm that selects the subset statistically closest to $\mathcal{\hat{Z}}_{T,\mathcal{C}}$ among $m$ random subsets, thereby improving the matching \textbf{effectiveness} and \textbf{stability}.

To select the final subset, we design a statistic evaluation metric $\mathcal{L}_{T,\mathcal{C}}$ to measure the distance between   the subset distribution  and  $\mathcal{\hat{Z}}_{T,\mathcal{C}}$, and select the subset with the minimal value. 
Specifically, $\mathcal{L}_{T,\mathcal{C}}$ computes the differences in the key statistics between two data distributions, including the mean, standard deviation, and skewness. 
$\mathcal{L}_{T,\mathcal{C}}$ can be expressed as the weighted sum of the three, 
\begin{equation}
\label{equ:L}
\begin{split}
    \mathcal{L}_{T,\mathcal{C}} &= \lambda_{\mu} \cdot \mathcal{L}_{\mu,T,\mathcal{C}} + \lambda_{\sigma} \cdot \mathcal{L}_{\sigma,T,\mathcal{C}} + \lambda_{\gamma_1} \cdot \mathcal{L}_{\gamma_1,T,\mathcal{C}},
\end{split}
\end{equation}
where $\mathcal{L}_{\mu,T,\mathcal{C}}$, $\mathcal{L}_{\sigma,T,\mathcal{C}}$ and $\mathcal{L}_{\gamma_1,T,\mathcal{C}}$  denotes the evaluation of $\mu_{T,\mathcal{C}}$,  $\sigma_{T,\mathcal{C}}$, and  $\gamma_{1,T,\mathcal{C}}$, respectively, with the formulation below, 
\begin{equation}
\scalebox{0.9}{$
    \mathcal{L}_{\mu,T,\mathcal{C}} = (\bar{z}_{T,\mathcal{C}} - \mu_{T,\mathcal{C}})^2 = (\frac{1}{n} \sum_{i=1}^{n} \hat{z}_{T,\mathcal{C}}^i - \mu_{T,\mathcal{C}})^2,
$}
\label{equ:kd}
\end{equation}
\begin{equation}
\scalebox{0.9}{$
\mathcal{L}_{\sigma,T,\mathcal{C}} = \left( \sqrt{\frac{1}{n} \sum_{i=1}^{n} (\hat{z}_{T,\mathcal{C}}^i - \mu_{T,\mathcal{C}})^2} - \sigma_{T,\mathcal{C}} \right)^2,
$}
\end{equation}
\begin{equation}
\scalebox{0.9}{$
    \mathcal{L}_{\gamma_1,T,\mathcal{C}} = \left(\frac{n}{(n-1)(n-2)} \sum_{i=1}^{n} \left(\frac{\hat{z}_{T,\mathcal{C}}^i - \mu_{T,\mathcal{C}}}{\sigma_{T,\mathcal{C}}}\right)^3-{\gamma_{1,T,\mathcal{C}}} \right)^2,
$
}
\label{equ:kd}
\end{equation}
where $\mu_{T,\mathcal{C}}, \sigma_{T,\mathcal{C}}$ and ${\gamma_{1,T,\mathcal{C}}}$ are the mean, standard deviation and skewness of $\mathcal{\hat{Z}}_{T,\mathcal{C}}$. 
$\hat{z}_{T,\mathcal{C}}^i$ indicates the $i^{th}$ element of the random subset.
As Gaussian distribution is perfectly symmetric, ${\gamma_{1,T,\mathcal{C}}}=0$.

With the above evaluation metric function, for the $k^{th}$ subset, we can compute its metric $\mathcal{L}^k_{T,\mathcal{C}}$. And we select  the $j^{th}$ subset with the smallest $\mathcal{L}_{T,\mathcal{C}}$ as the most representative subset:
\begin{equation}
\begin{split}
   j = \underset{1 \leq k \leq {m}}{\arg\min}(\mathcal{L}^k_{T,\mathcal{C}}).
\end{split}
\label{equ:kd}
\end{equation}
After we select the most representative subset, they are converted to real images through \cref{equ:ddim}
with DDIM and a VAE decoder. 

\begin{figure}[t]
    \captionsetup[subfloat]{captionskip=-8pt}
    \centering
    \vspace{-2em}
     \hfill
    \subfloat[][Base-DDPM]
    {\includegraphics[width=0.4\linewidth]{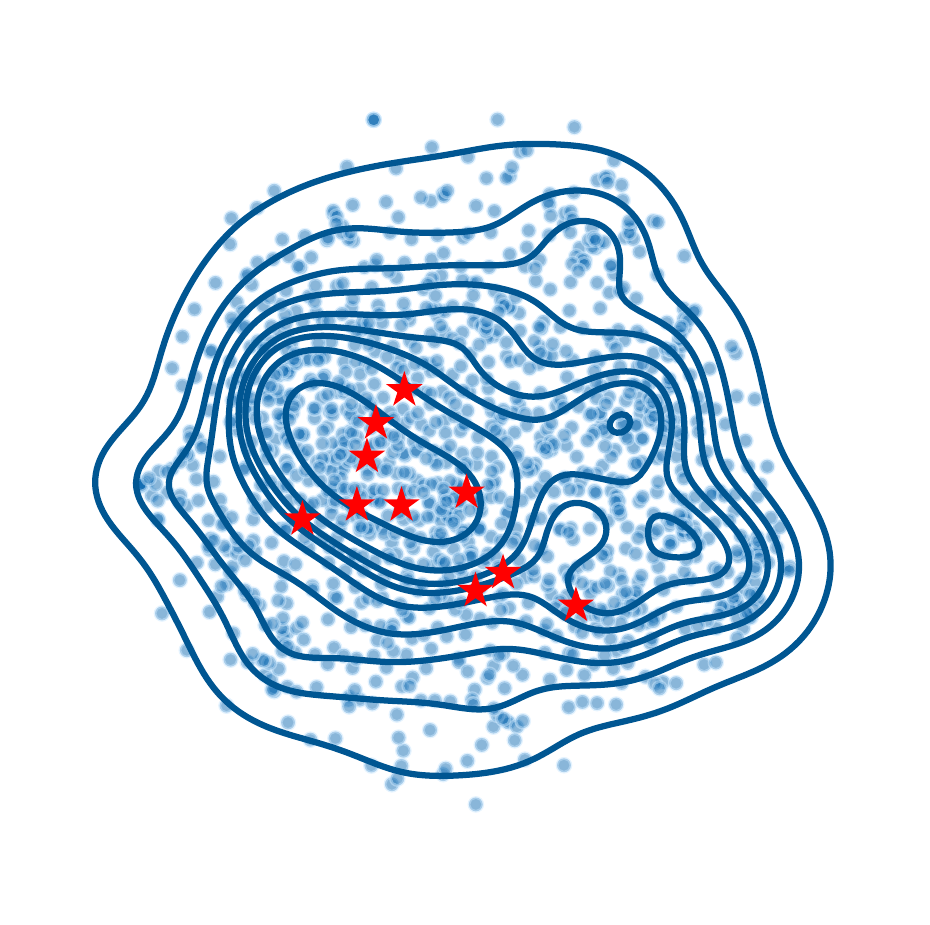}\label{fig:ddpm-space-normality}}
    \hfill
    \subfloat[][\methodAbbr]
    {\includegraphics[width=0.4\linewidth]{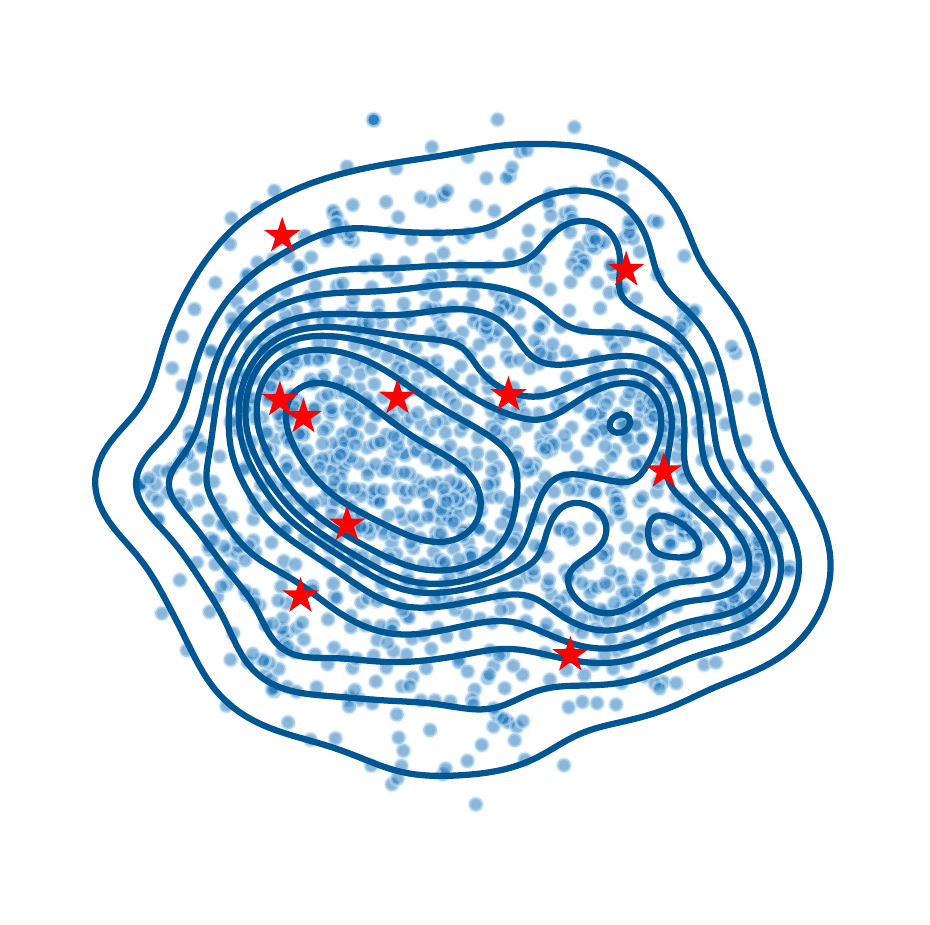}\label{fig:ours-space-normality}}
    \hfill\null
    \vspace{-.6em}
    \caption{\textbf{t-SNE visualization of the representative latents in VAE space generated by \cref{equ:ddpm} and \methodAbbr for class “Goldfish".} 
    It can be observed that the latents accurately represent the VAE distribution in \methodAbbr.}
    \label{fig:vae-representative}
\end{figure}

\subsection{Advantages of \methodAbbr}\label{subsec:strengths}
Compared to other methods, we highlight our approach has the following notable advantages: 
\begin{enumerate*}[label=(\arabic*)]
    \item We present a domain mapping method to map VAE latent domain into Gaussian domain with \textbf{better normality}, 
    leading to  more \textbf{accurate} and \textbf{efficient}  distribution matching, as illustrated in \cref{fig:noise-space-normality}.
    \item We ensure that \textbf{the entire subset is representative} closely aligned with the desired distribution, rather than just sampling individual elements without considering their overall distribution.
    \item 
    Our sampling algorithm is \textbf{both time- and space-efficient}. 
    The sampling is performed based on the statistic parameters and we can sample different subsets in parallel on the GPU.
    Our sampling implementation only takes 2.6 seconds per class on a single RTX A6000 with ${m}=1e6$ for ImageNet-1K, $\text{IPC}=10$. Additional runtime results under different settings are provided in Appendix B-\cref{tab:A1001e5,tab:A1001e6,tab:A60001e5}. Moreover, instead of storing representative latents, we can generate the distilled image directly from statistical parameters, which are space-efficient as shown in Appendix B-\cref{fig:storage}.
\end{enumerate*}

\section{Main Results}
\subsection{Experimental Details}
\noindent \textbf{Datasets}.~
Experiments are conducted on both small-scale and large-scale datasets.
For small-scale datasets, we use  CIFAR-10 and CIFAR-100 \cite{krizhevsky2009learning} with  $32  \times 32$  resolution. 
%
%
For large-scale datasets, we use Tiny-ImageNet \cite{le2015tiny} with $200$ classes ($500$ images per class, 64 $\times$ 64 size) and ImageNet-1K \cite{deng2009imagenet} with $1,000$ classes ($1$M images, 224 $\times$ 224 resolution).

\noindent \textbf{Network architectures}.~
To validate the applicability of our method across different architectures, we adopt ResNet-18/ResNet-101 \cite{he2016deep}, MobileNet-V2 \cite{sandler2018mobilenetv2}, and  VGG-11 \cite{simonyan2014very} as backbones, following prior works \cite{du2024diversity,yin2023squeeze,sun2024diversity}.

\noindent \textbf{Baselines}.
We compare \methodAbbr with four state-of-the-art methods: SRe\textsuperscript{2}L~\cite{yin2023squeeze}, DWA~\cite{du2024diversity}, D\textsuperscript{4}M~\cite{Su_2024_CVPR}, RDED~\cite{sun2024diversity}, following the same evaluation configuration.
We exclude comparison with Minimax \cite{gu2024efficient} in the main \cref{tab:performance_comparison}, since their method focuses on handling small subsets of ImageNet-1K and requires training multiple diffusion models with expensive computation for large-scale scenarios. Instead, we report results under their setup in \cref{subsec:compare Minimax}.
%
\begin{table}[t]
\centering
\footnotesize
\renewcommand{\arraystretch}{1.0}
\resizebox{0.5\textwidth}{!}{
\begin{tabular}{lc|ccc}
\toprule
\multicolumn{2}{c|}{Student\textbackslash Teacher} & ResNet-18 & MobileNet-V2 & VGG-11 \\
\midrule
\multirow{3}{*}{ResNet-18} & SRe$^2$L & $21.3 \pm 0.6$ & $15.4 \pm 0.2$ & -\\
 & RDED & $42.3 \pm 0.6$ & $40.4 \pm 0.1$ & $36.6 \pm 0.1$ \\
 & 
 \cellcolor{LightCyan}Ours & 
 \cellcolor{LightCyan}$\textbf{44.3} \pm \textbf{0.3}$ & 
 \cellcolor{LightCyan}$\textbf{42.3} \pm \textbf{0.7}$ & 
 \cellcolor{LightCyan}$\textbf{38.3} \pm \textbf{0.2}$ \\
\midrule
\multirow{3}{*}{MobileNet-V2} & SRe$^2$L & $19.7 \pm 0.1$ & $10.2 \pm 2.6$ & - \\
& RDED & $34.4 \pm 0.2$ & $33.8 \pm 0.6$ & $28.7 \pm 0.2$ \\
 & 
 \cellcolor{LightCyan}Ours &
 \cellcolor{LightCyan}$\textbf{43.4} \pm \textbf{0.3}$ &
 \cellcolor{LightCyan}$\textbf{46.4} \pm \textbf{0.2}$ &
 \cellcolor{LightCyan}$\textbf{37.8} \pm \textbf{0.4}$ \\
\midrule
\multirow{3}{*}{VGG-11} & SRe$^2$L & $16.5\pm0.1$ & $10.6\pm0.1$ & - \\
& RDED & $22.7 \pm 0.1$ & $21.6 \pm0.2$ & $23.5 \pm 0.3$ \\
& 
\cellcolor{LightCyan}Ours & \cellcolor{LightCyan}$\textbf{25.7} \pm \textbf{0.4}$ &
\cellcolor{LightCyan}$\textbf{24.8} \pm \textbf{0.4}$ &
\cellcolor{LightCyan}$\textbf{28.1} \pm \textbf{0.1}$ \\
\bottomrule
\end{tabular}
}
\caption{\textbf{Evaluating Top-1 accuracy for cross-architecture generalization on ImageNet-1K}, $\text{IPC}=10$. 
As VGG-11 lacks BN layers,  the results of SRe$^2$L  
are not available with `-'.}
\label{tab:cross}
\end{table}
\noindent \textbf{Implementation details}.~
We adopt the pre-trained Diffusion Transformer~(DiT) and VAE from~\citet{peebles2023scalable} in our framework, originally trained on ImageNet-1K.
We further adjust the conditioning labels of DiT, and fine-tune the pre-trained model with $400$ epochs for each dataset (Tiny-ImageNet, CIFAR-10 and CIFAR-100) to adapt its generative capacity to the specific data distributions.
For distillation, we employ $31$ steps for DDIM inversion and sampling.
During the validation, we follow other works \cite{yin2023squeeze,du2024diversity,Su_2024_CVPR,sun2024diversity} to use soft-label of the teacher model as supervision for training.
All experiments are conducted on Nvidia RTX A6000 GPUs or Nvidia A100 40GB GPUs.

\subsection{Comparison with State-of-the-art Methods}
As shown in \cref{tab:performance_comparison}, \textbf{\methodAbbr demonstrates superior performance across all IPCs compared with  baselines}.

\noindent \textbf{Large-scale datasets}.~
We first validate the practicality of \methodAbbr on Tiny-ImageNet and ImageNet-1K at various IPCs.
For Tiny-ImageNet, while RDED performs well on ResNet-18, its reliance on the teacher model leads to a significant drop on cross ResNet-101. 
In contrast, \methodAbbr consistently delivers SOTA performance on both ResNet-18 and  ResNet-101.
Similarly, for ImageNet-1K, \methodAbbr  achieves higher performance across all IPCs compared with all baselines.
\noindent \textbf{Small-scale datasets}.~
For CIFAR-10 and CIFAR-100, \methodAbbr surpasses all baselines. Notably, compared with the best-performing baseline RDED, on CIFAR-10, our method achieves significant improvements of $12.5\%$ for ResNet-18 and $17.4\%$ for  ResNet-101 at $50$ IPC. 

In addition, we provide a comparison with other state-of-the-art methods that use validation settings different from ours. To ensure fairness, we evaluate our method under their settings, as shown in \cref{tab:more_com_cifar,tab:more_com_img}, to demonstrate our superiority.

\subsection{Cross-architecture Generalization} \label{subsec:cross-architecture}
To further evaluate cross-model performance,  on ImageNet-1K, we compare \methodAbbr with the recent baseline SRe\textsuperscript{2}L and RDED (SOTA performance on ImageNet-1K). 
For \methodAbbr, the teacher model refers to using its soft label for validation.
As depicted in \cref{tab:cross}, we achieve a significant accuracy improvement   across all cross-model evaluations.
We highlight that with \methodAbbr, \textbf{a one-time cost is sufficient to achieve satisfactory results across various models}. Other baselines need to run their algorithms multiple times to generate multiple distilled datasets  when the model architecture changes. 
The comparison results for more architectures are provided in Appendix B-\cref{tab:results}.

\section{Analysis}

\subsection{Ablation Studies} \label{subsec:DM}

\noindent \textbf{Effectiveness of Domain Mapping by DDIM Inversion}.~
To demonstrate the effectiveness of domain mapping with DDIM inversion, we experiment with two configurations: (i) domain mapping with \cref{equ:ddpm}  (denoted as Base-DDPM), and (ii) domain mapping with \cref{equ:ddim_inversion}  (denoted as Base-RS). Note that for both configurations, we perform distribution matching and individual Gaussian sampling (i.e., random sampling (RS)), while our proposed group sampling method in Section~\ref{subsec:gaussian sampling} is not applied. 

As shown in Table~\ref{tab:ablation_study}, by comparing the results of Base-DDPM and Base-RS, we can observe that   domain mapping with DDIM inversion  outperforms that of DDPM with an accuracy improvement of $11.5\%$. 
As we discussed in \cref{subsec:domain mapping}, the added noise through DDPM  causes     structural information loss and domain shift for the mapping from $\mathcal{Z}_{0,\mathcal{C}}$ to $\mathcal{Z}_{T,\mathcal{C}}$, leading to difficulties to   obtain a representative subset for $\mathcal{Z}_{0,\mathcal{C}}$. 
By contrast, the determinism of DDIM inversion resolves the issue with  its information preservation and structural consistency, as illustrated in \cref{fig:ours-space-normality}.

\begingroup
\begin{table}[t]
    \centering
    \renewcommand{\arraystretch}{1.0}
    \footnotesize
    \begin{tabular}{l|c}
        \toprule
        DM / Sampling Design & Acc.~(\%)\\ 
        \midrule
        Base-DDPM & $37.3\pm0.7$ \\ 
        Base-RS  & $41.6\pm0.2 $ \\ 
        $\mathcal{L}_\mu$     & $42.6\pm0.1$ \\
        $\mathcal{L}_\sigma$     & $42.3\pm0.2$ \\ 
        $\mathcal{L}_{\gamma_1}$ & $42.4\pm0.1$ \\ $\mathcal{L}_\mu+\mathcal{L}_\sigma$      
        & $43.3\pm0.2$ \\ $\mathcal{L}_\mu+\mathcal{L}_{\gamma_1}$     
        & $43.0\pm0.1$ \\ $\mathcal{L}_\sigma+\mathcal{L}_{\gamma_1}$    
        & $42.5\pm0.2$ \\ \rowcolor{LightCyan}\textbf{$\mathcal{L}_\mu+\mathcal{L}_\sigma+\mathcal{L}_{\gamma_1}$ (Ours)}& $\bf{44.3}\pm\bf{0.3}$ \\ 
 \bottomrule
    \end{tabular}
    \caption{\textbf{Ablations of  \methodAbbr on ResNet-18 for ImageNet-1K with $\text{IPC}=10$.} `DM' indicates domain mapping, and `RS' indicates random sampling. } 
    \label{tab:ablation_study}
\end{table}

\endgroup
%
%
%
%
%
%
\begin{figure}[t]
    \centering   \includegraphics[width=\linewidth]{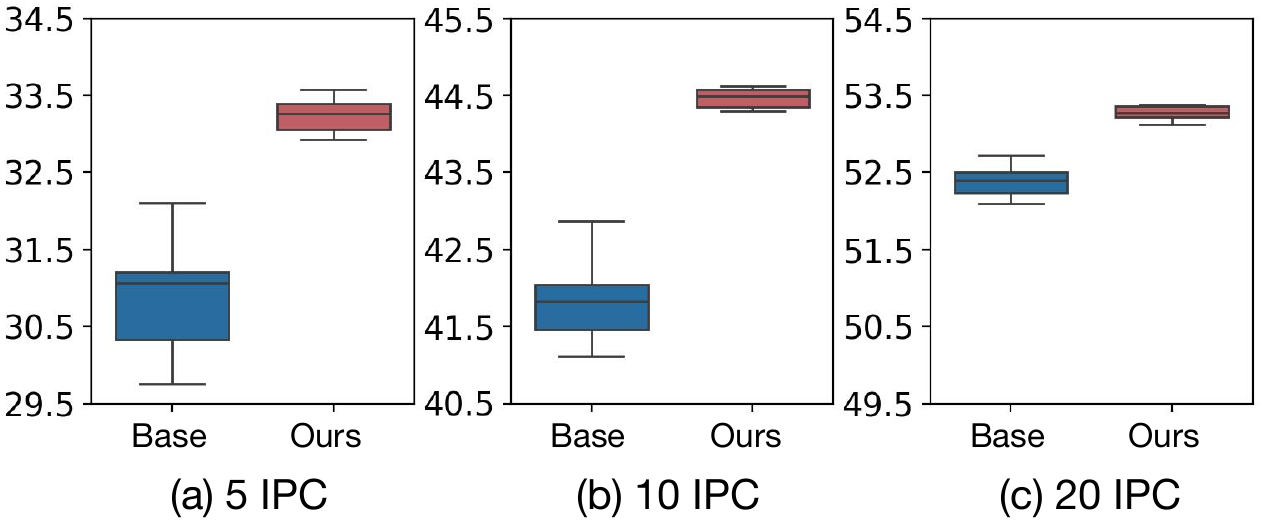}
    \caption{\textbf{Validation of the effectiveness and stability of   group sampling  on ImageNet-1K.} For  our \methodAbbr and Base-RS, we generate 3 distilled datasets at each IPC. Each dataset undergoes 3 rounds of validation, resulting in  9 data points   per box plot.}
    \label{fig:boxplot}
    \vspace{-0.1cm}
\end{figure}

\noindent \textbf{Effectiveness of Group Sampling}.~
As shown in Table~\ref{tab:ablation_study}, by comparing our results with Base-RS,   our proposed group sampling method can significantly outperform individual Gaussian sampling, demonstrating its effectiveness. 

Furthermore, we conduct ablation studies with different combinations in~\cref{equ:L} to verify the contribution of each metric in the sampling schedule. 
As demonstrated in \cref{tab:ablation_study}, each metric individually increases the accuracy, indicating that the subset selection effectively brings the distribution of the sampled representative subset closer to $\mathcal{\hat{Z}}_{T,\mathcal{C}}$.
Combining all the metrics can lead to the best performance with the most representative subset. 

%
Besides, we highlight that our group sampling method enhances the stability of the distilled dataset.
As presented in \cref{fig:boxplot}, we generate  distilled datasets for the same dataset multiple times with both our \methodAbbr and  Base-RS.
%
The results demonstrate that \methodAbbr   not only improves the accuracy  but also enhances stability with smaller variance.

\subsection{Analysis of Different Inversion Steps}  \label{subsec:different_steps}
In \cref{lemma:ddim}, as the number of  inversion steps $t$ increase, $\mathcal{\hat{Z}}_{t,\mathcal{C}}$ gradually transits from a complex Gaussian mixture distribution to a standard Gaussian distribution, 
leading to a easier distribution matching.
%
We provide feature visualizations to illustrate the changes as $t$ increases in Appendix B-\cref{fig:5steps}, validating \cref{lemma:ddim}.
However, as $t$ increase, the added noise also increases, gradually diminishing the retention of original image structural information and reducing the reconstruction quality of DDIM inversion.

\begin{figure}[t]
    \centering
\includegraphics[width=0.9\linewidth]{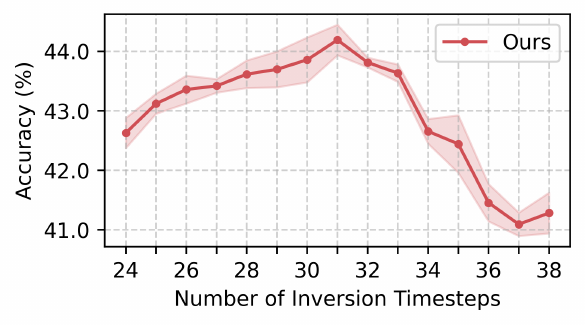}
\vspace{-.8em}
    \caption{\textbf{The accuracy variation under different inversion timesteps for ImageNet-1K}, $\text{IPC}=10$.}
    \label{fig:effect of timesteps}
\end{figure}

Specifically, there is a trade-off between maintaining the Gaussian assumption and preserving image structural information across different steps $t$. 
To assess the impact of different steps, we conduct experiments with varying $t$.  
As shown in \cref{fig:effect of timesteps}, the accuracy initially improves with increasing $t$, reaches a peak at $t=31$, and then starts to decline.
When $t$ is small (e.g., $t=20$), the distribution is a mixture of Gaussians, and our distribution matching with a single Gaussian in \cref{lemma:ddim} is not able to accurately describe the Gaussian mixtures, leading to certain performance loss. When $t$ becomes large (e.g., $t=40$), although our distribution matching can accurately represent the real distributions which becomes more normal, the real distributions suffer from more significantly structural information loss due to adding more noise, which in turn degrades the performance of DDIM inversion.

\begin{figure}[t]
    \centering   \includegraphics[width=0.9\linewidth]{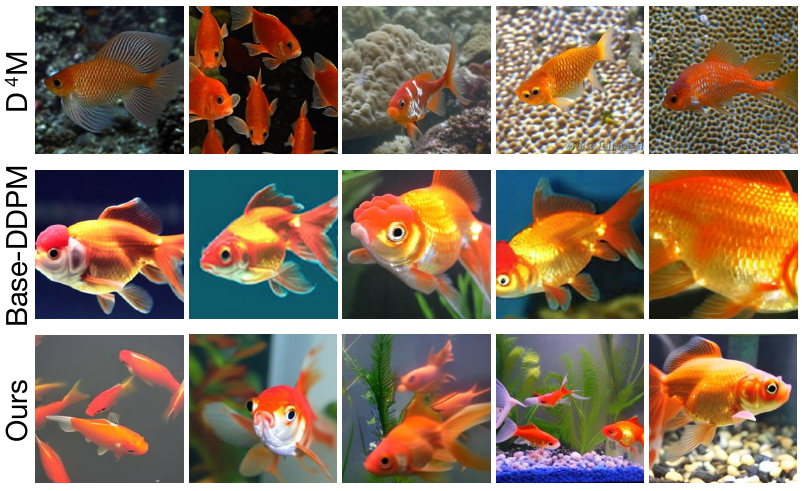}
    \caption{\textbf{Visualization of the distilled images for the class ``goldfish" on ImageNet-1K}, $\text{IPC}=5$.}  
    \label{fig:5ipc}
\end{figure}
\begin{table}[t]
\centering
\resizebox{\linewidth}{!}{
\begin{tabular}{llcccc}
\toprule
\multirow{2}{*}{Dataset}& \multirow{2}{*}{IPC} & \multicolumn{2}{c}{ResNet-18} & \multicolumn{2}{c}{ResNet-101} \\
\cmidrule(lr){3-4} 
\cmidrule(lr){5-6} 
& & D3S & Ours & D3S & Ours \\
\midrule
\multirow{2}{*}{ImageNet-1K} 
    & 10 & $39.1 \pm 0.3$ 
         & $\textbf{44.8} \pm \textbf{0.1}$ 
         & $42.1 \pm 3.8$
         & $\textbf{52.8} \pm \textbf{0.6}$\\
    & 50 & $60.2 \pm 0.1$ 
         & $\textbf{60.2} \pm \textbf{0.0}$
         & $65.3 \pm 0.5$ 
         & $\textbf{66.8} \pm \textbf{0.1}$\\
\bottomrule
\end{tabular}
}
\caption{\textbf{Comparison with D3S \cite{loolarge} on ImageNet-1K.} Following the default setup of D3S with $5$ pre-trained teacher models, our \methodAbbr apply soft labels  from these $5$ teacher models.}
\label{tab:vsD3S}
\vspace{-1mm}
\end{table}
\begin{table}[h]
\centering
\resizebox{0.4\textwidth}{!}{
\begin{tabular}{cc|cc}
\toprule
IPC & Model & Minimax & Ours \\
\midrule
\multirow{2}{*}{10} 
  & ResNet-18    & $37.6 \pm 0.9$   & $\textbf{39.6} \pm \textbf{1.0}$     \\
  & ResNetAP-10 & $39.2 \pm 1.3$   & $\textbf{40.7} \pm \textbf{1.0}$    \\
\midrule
\multirow{2}{*}{50} 
  & ResNet-18    & $57.1 \pm 0.6$   & $\textbf{57.6} \pm \textbf{0.4}$   \\
  & ResNetAP-10 & $56.3 \pm 1.0$   & $\textbf{59.3} \pm \textbf{0.4}$   \\
\midrule
\multirow{2}{*}{100} 
 & ResNet-18    & $65.7 \pm 0.4$   & $\textbf{66.8} \pm \textbf{0.6}$   \\
 & ResNetAP-10 & $64.5 \pm 0.2$   & $\textbf{64.7} \pm \textbf{0.3}$   \\
\bottomrule
\end{tabular}
}
\caption{\textbf{Comparison with Minimax \cite{gu2024efficient} under hard labels across different models and IPCs.}}
\label{tab:com_minimax}
\end{table}

%

\subsection{Image Visualization}
We present the visualization results of distilled images generated by D4M, Base-DDPM, 
and our \methodAbbr.
%
\cref{fig:5ipc} shows that the images generated by D$^4$M \cite{Su_2024_CVPR} are simplistic and unrepresentative due to the incorrect retrieval of representative latents.
%
For Base-DDPM with \cref{equ:ddpm}, 
the noise space fails to generate representative latents due to randomness, resulting in outputs with overly simplistic structures, corresponding to \cref{fig:ddpm-space-normality}.
In contrast, the images generated by \methodAbbr are both representative and diverse. 

\subsection{\methodAbbr Outperforms Under Hard Labels}\label{subsec:compare Minimax}
We give the comparison results with Minimax \cite{gu2024efficient}, which is the state-of-the-art method with hard labels under their main evaluation setting on ImageWoof \cite{howard2019smaller}. As demonstrated in \cref{tab:com_minimax}, \methodAbbr with $224\times224$ resolution outperforms Minimax \cite{gu2024efficient} with $256\times256$ resolution.

\subsection{Improving Results with More Soft Labels}
D3S \cite{loolarge} proposes to simultaneously map the distribution of multiple teacher models. 
During evaluation, the soft labels from multiple teacher models are averaged to supervise the training of the student model.
The drawback is the increased costs to train multiple different teachers. 
%

Although \methodAbbr does not rely on teacher models during the distillation, we find that using multiple soft labels during evaluation can further improve the  performance.
We present \methodAbbr and D3S with $5$ teacher models following  their default setting on ImageNet-1K in \cref{tab:vsD3S}.
Since \methodAbbr with one soft label outperforms D3S with five on TinyImageNet across various IPCs, we do not experiment with TinyImageNet. 
%
As shown in \cref{tab:performance_comparison} and \cref{tab:vsD3S}, leveraging $5$ soft labels yields slight improvements over $1$ soft labels across different IPCs and ours consistently outperforms D3S.
For $10$ IPC, \methodAbbr with $1$ soft label outperforms D3S with $5$ soft labels on  ImageNet-1K.
%

\subsection{Robustness of \methodAbbr}     \label{subsec:Robustness}
As shown in \cref{alg:base method}, \methodAbbr~involves inherent randomness: initializing different seeds for the random sampling of  multiple  subsets.  
%
%
For the diffusion-based   D\textsuperscript{4}M~\cite{Su_2024_CVPR}, the randomness arises from two main sources: (i) Different seeds result in varying initial clusters in the K-means algorithm, and (ii) Different seeds generate varying initial noise by \cref{equ:ddpm} for the distilled dataset generation.

\begin{figure}[t]
    \centering   \includegraphics[width=0.9\linewidth]{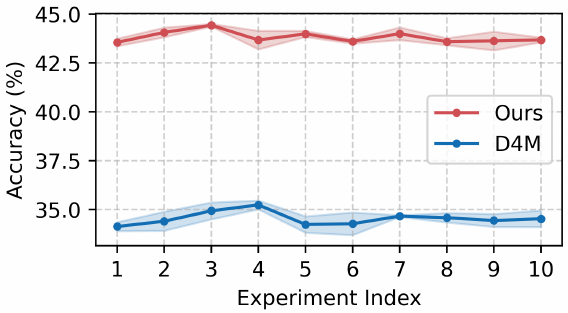}
    \caption{\textbf{The robustness of \methodAbbr across different seeds on TinyImageNet.} $\text{IPC}=10$. The ``Experiment Index" represents ten different seeds selected at random.}
    \label{fig:vsD4M}
\end{figure}

To validate the consistent superiority across various seeds of \methodAbbr, we conduct both our \methodAbbr and D\textsuperscript{4}M~\cite{Su_2024_CVPR} ten times with different seeds.
As shown in \cref{fig:vsD4M}, our \methodAbbr outperforms D\textsuperscript{4}M~\cite{Su_2024_CVPR} by approximately $27.5\%$ across various seeds.

\subsection{Storage Requirements Smaller than $D$}    \label{subsec:Storage}
For \methodAbbr, all the data information is encompassed in $\mathcal{\hat{Z}}_T$. Therefore, we highlight that by storing only the statistical parameters (mean and variance) of $\mathcal{\hat{Z}}_T$ and the DiT pretrained weights of the diffusion model, we can preserve all the details of the distilled datasets under any IPCs. 
As shown in Appendix B-\cref{fig:storage}, it further reduces the required storage space compared to $\mathcal{D}$.
This approach is particularly effective for large datasets, resulting in a substantial reduction in storage requirements.

\section{Conclusion}
In this work, we thoroughly examine the challenges in existing diffusion-based dataset distillation methods, such as inaccurate distribution matching, distribution deviation with random noise, and the reliance on separate sampling.
Based on this, we introduce a novel diffusion-based framework, \methodAbbr, to generate a highly representative distilled dataset. 
Our method achieves state-of-the-art validation performance on various datasets for different models.

\section*{Impact Statement}
This paper presents work whose goal is to advance the field of Machine Learning. There are many potential societal consequences of our work, none which we feel must be specifically highlighted here.

\section*{Acknowledgments}
This work is partially supported by the National Science Foundation under Award IIS-2310254. We would like to express our sincere gratitude to the reviewers for their invaluable feedback and constructive comments to improve
the paper.

\bibliography{reference}
\bibliographystyle{icml2025}

\newpage
\appendix
\onecolumn

\appendix

\setcounter{table}{0}
\renewcommand{\thetable}{A\arabic{table}}
\renewcommand*{\theHtable}{\thetable}

\setcounter{figure}{0}
\renewcommand{\thefigure}{A\arabic{figure}}
\renewcommand*{\theHfigure}{\thefigure}

\begin{center}
    \noindent\textbf{\huge Appendix}
\end{center}

In the appendix, we include additional details that were omitted in the main text:
\begin{itemize}[leftmargin=10pt]
    \item Section \crefname{appendix}{}{}\cref{sec:Theoretical}: Theoretical Analysis.
    \item Section \crefname{appendix}{}{}\cref{sec:ad_implementation}: More Implementation Details.
    \item Section \crefname{appendix}{}{}\cref{sec:ad_Experiments}: Additional Experiments.
    \item Section \crefname{appendix}{}{}\cref{sec:ad_Visualization}: Image Visualization.
\end{itemize}

\section{Theoretical Analysis} \label{sec:Theoretical}

\begin{lemma}
Each latent in VAE latent space is randomly sampled from a distinct component of a multi-component Gaussian mixture distribution; hence, the VAE latent is hard-to-fit.
\end{lemma}
%
\textit{Proof.} The Variational Autoencoder (VAE) \cite{kingma2019introduction} learns the latent distribution $\mathcal{Z} = \{z_i | i=1,...,M\}$ of input images $\mathcal{X} = \{x_i | i=1,...,M\}$ through probabilistic modeling and is capable of sampling from this distribution to generate new images.
In training, for each image $x_i$, VAE aims to maximize the marginal likelihood $p(x_i)$. The loss function can be defined by minimizing the negative evidence lower bound:
\begin{equation}
\mathcal{L}_i = -\mathbb{E}_{q(z_i|x_i)} \big[ \log p(x_i|z_i) \big] + \text{KL}\big(q(z_i|x_i) \| p(z)\big).
\end{equation}
The first term represents the reconstruction loss of $x_i$. The second term is the KL divergence loss, which is used to measure the difference between the latent distribution $q(z_i|x_i)$ produced by the encoder and the prior distribution $p(z)\sim\mathcal{N}(0,I)$.
The output of the VAE encoder for $x_i$ is defined as a Gaussian distribution $q(z_i|x_i)\sim\mathcal{N}(\mu_i,\sigma_i^2)$, where $\mu_i$ and $\sigma_i^2$ are the mean and variance learned by the encoder. The latent $z_i$ is then obtained by randomly sampling from this Gaussian distribution.
Therefore, for each class $\mathcal{C}$ containing $\mathbf{m}$ images, $\mathcal{Z}_\mathcal{C}$ is modeled as a discrete distribution, where the $\mathbf{m}$ latents are independently sampled from $\mathbf{m}$ distinct Gaussian distributions.
The probability density function of the multi-component Gaussian mixture distribution can be expressed as:
\begin{equation}
p(z) = \sum_{i=1}^{\mathbf{m}} \alpha_i \cdot \mathcal{N}(x \mid \mu_i, \Sigma_i),
\end{equation}
where \(\mathcal{N}(x \mid \mu_i, \Sigma_i)\) denotes the \(i\)-th Gaussian distribution, with \(\mu_i\) and \(\Sigma_i\) representing its mean and covariance, respectively. The weights \(\alpha_i\) satisfy the following condition: $\sum_{i=1}^{\mathbf{m}} \alpha_i = 1$.
Since the position of each Gaussian component of this Gaussian mixture distribution is random, it results in a complex and hard-to-fit distribution.

\begin{lemma}
\label{pro:ddim}
For DDIM inversion with $T'$ steps, when $t \; (0\leq t\leq T')$ is sufficiently large, $\mathcal{\hat{Z}}_{t,\mathcal{C}}$ can be approximated as a Gaussian distribution.
\end{lemma}
\textit{Proof.} In DDIM training, the forward process also follows \cref{equ:ddpm}, which implies that $p(x_{T'})\sim\mathcal{N}(0,I)$. 
During inference, the process from $x_T$ to $x_0$ is deterministic, following \cref{equ:ddim}.
Since $\alpha_{t}$ is a pre-defined hyperparameter for each $t$, the process from $z_t$ to $z_{t-1}$ can be viewed as a linear transformation: 
\begin{equation}
z_{t-1} = v_t \cdot z_t + w_t, \quad \text{where} \quad v_t = \sqrt{\frac{\alpha_{t-1}}{\alpha_{t}}}, \quad w_t = \sqrt{\alpha_{t-1}} \left( \sqrt{\frac{1}{\alpha_{t-1}} - 1} - \sqrt{\frac{1}{\alpha_{t}} - 1} \right) \varepsilon_{\theta}(z_t, t, \mathcal{C}).
\label{equ:12}
\end{equation}
For different sampled latent $z_t^i$, $v_t^i$ remains the same, while $w_t^i$ changes because the network $\varepsilon_{\theta}$ produces different outputs for each $z_t^i$.

We approach from the DDIM sampling, which is the reverse of the DDIM inversion, proceeding from $T'$ to $0$.
Specifically, consider the deterministic transformation from the set $\mathcal{Z}_{T',\mathcal{C}} = \{z_{T'}^i | i = 1, \dots, \mathbf{m} \}$ to $\mathcal{Z}_{T'-1,\mathcal{C}}$ (each noise in $\mathcal{Z}_{T',\mathcal{C}}$ corresponds one-to-one with each latent in $\mathcal{Z}_{0,\mathcal{C}}$). Each latent in $\mathcal{Z}_{T'-1,\mathcal{C}}$ is a latent sampled from the Gaussian distribution:
\begin{equation}
z_{T'-1}^i \sim \mathcal{N}(w_{T'}^i, (v_{T'})^2).
\end{equation}

According to the law of large numbers~\cite{hsu1947complete}, the latents in the discrete distribution $\mathcal{Z}_{T'-1,\mathcal{C}}$ can be interpreted as i.i.d. latents from  continuous distribution $\mathcal{\hat{Z}}_{T'-1,\mathcal{C}}$. Thus, the $\mathcal{\hat{Z}}_{T'-1,\mathcal{C}}$ is a Gaussian mixture distribution. 

Next, consider the transformation from $\mathcal{Z}_{T'-1,\mathcal{C}}$ to $\mathcal{Z}_{T'-2,\mathcal{C}}$. Each latent in $\mathcal{Z}_{T'-2,\mathcal{C}}$ is sampled from:
\begin{equation}
z_{T'-2}^i \sim \mathcal{N}(v_{T'-1} \cdot w_{T'}^i + w_{T'-1}^i, (v_{T'-1})^2 \times (v_{T'})^2).
\end{equation}

Based on above iteration, the transformation from $\mathcal{Z}_{t+1,\mathcal{C}}$ to $\mathcal{Z}_{t,\mathcal{C}}$ can be generalized as follows:  
\begin{equation}
z_{t}^i \sim \mathcal{N}\left(\sum_{k=t+1}^{T'} \left( \prod_{j=k+1}^{T'} v_{T'+t-j+1} \right) w_{T'+t-k+1}^i, \prod_{j=t+1}^{T'} (v_j)^2\right),
\end{equation}
Therefore, $\mathcal{\hat{Z}}_{t,\mathcal{C}}$ ($0 \leq t \leq T'-1$) is a Gaussian mixture distribution.

Now, in the DDIM sampling process, we have the recurrence relation for $\alpha_t$:
\begin{equation}
\alpha_t = \prod_{i=1}^t \alpha_i', \quad \text{where} \quad \alpha_i' < 1 \& \alpha_i' \to 1,
\end{equation}

this implies the following formulas:
\begin{equation}
|\alpha_t - \alpha_{t-1}| < \epsilon, \quad \text{where} \quad \epsilon \text{ is a small positive constant,}
\label{equ:16}
\end{equation}
\begin{equation}
\alpha_t \to 0 \quad \text{as} \quad t \to T'.
\label{equ:17}
\end{equation}

When $t$ is relatively large, on one hand, as indicated by \cref{equ:12,equ:16,equ:17}, the effect of $\varepsilon_{\theta}(z_t, t, \mathcal{C})$ on $w_t$ becomes minimal; on the other hand, since the number of iterations is small, the change in the statistical properties of the original standard Gaussian distribution is not particularly noticeable. Hence, the distributions of the Gaussian components in $\mathcal{\hat{Z}}_{t,\mathcal{C}}$ are not significantly different, and $\mathcal{\hat{Z}}_{t,\mathcal{C}}$ can be well-approximated by a Gaussian distribution.


\section{More Implementation Details}    \label{sec:ad_implementation}
For the group sampling, we set $m=100,000$ for the ImageNet-1K, $m=1,000,000$ for the TinyImageNet, and $m=5,000,000$ for CIFAR-10, CIFAR-100.
The hyper-parameters $\lambda_{\mu}$, $\lambda_{\sigma}$, $\lambda_{\gamma_1}$ are set to $1, 1, 0.5$, respectively.
Regarding the computational cost of DDIM inversion, our method only requires approximately 4.5 hours on a single node with 8 A100 40G GPUs on ImageNet-1K.

For validation, the parameter settings vary slightly across methods. We adhere to the configurations in \cite{sun2024diversity}, as detailed in \cref{tab:valid_setting}.
For other methods, we primarily use the results reported in their paper. If a relevant experiment is unavailable, we generate the distilled dataset using their code and validate it under the same settings as ours.

\begin{table}[H]
\centering
\label{tab:hyperparameters}
\begin{tabular}{lcccc}
\toprule
Parameter & CIFAR-10 & CIFAR-100 & Tiny-ImageNet & ImageNet-1K \\
\midrule
Optimizer & \multicolumn{4}{c}{AdamW} \\
Learning Rate & \multicolumn{4}{c}{0.01} \\
Weight Decay & \multicolumn{4}{c}{0.01} \\
Batch Size & \multicolumn{4}{c}{128} \\
Augmentation & \multicolumn{4}{c}{RandomResizedCrop + Horizontal Flip} \\
LR Scheduler & \multicolumn{4}{c}{CosineAnneal} \\
Tempreture & \multicolumn{4}{c}{20} \\
Epochs &400 & 400 & 300 & 300 \\
\bottomrule
\end{tabular}
\caption{\textbf{Hyper-parameter setttings used for our validation.}}
\label{tab:valid_setting}
\end{table}

\section{Additional Experiments}
\label{sec:ad_Experiments}
\subsection{Validate the Limitations of D$^4$M}
\begin{figure}[H]
    \centering
    \includegraphics[width=0.6\linewidth]{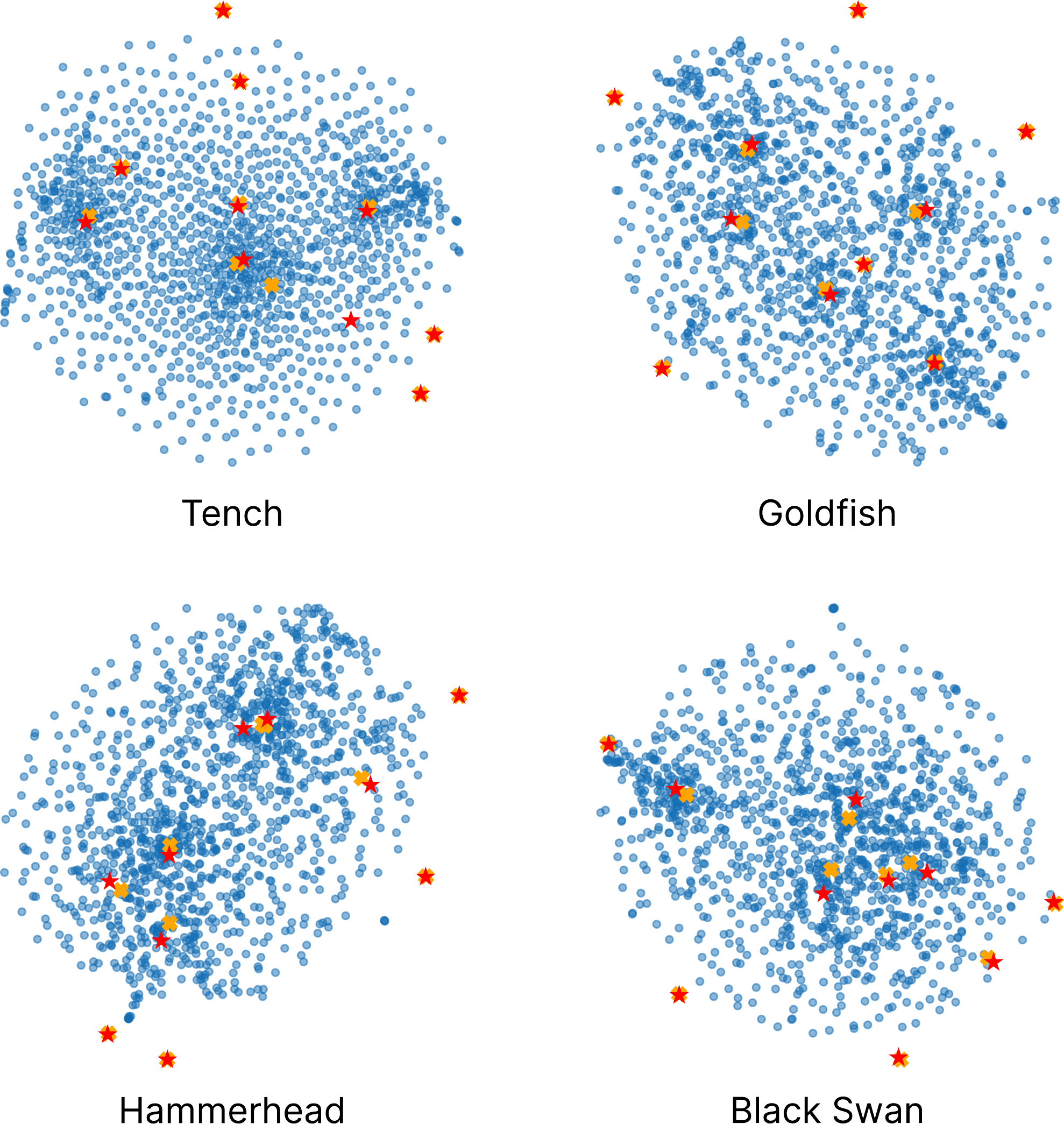}
    \caption{\textbf{Illustration of representation-shift in D\textsuperscript{4}M for different classes, ImageNet-1K, $IPC=10$.} The blue dots indicate the latents in the VAE space. \textcolor{orange}{\ding{56}} represent the cluster centers after the VAE encoding. \red{\ding{72}} denote the latents after adding noises and removing noises for decoding the distilled images.}
    \label{fig:d4m-shift}
\end{figure}

We use D$^4$M as an example here to validate the validity of our analysis in \cref{sec:motivation} by visualizing the latents in the VAE space.
As illustrated in \cref{fig:d4m-shift}, we visualize the cluster centers in the VAE encoding space, which are treated as representative latents of the method, and marked as \textcolor{orange}{\ding{56}}. 
It is evident that there are many outliers in the VAE space, and the K-means algorithm treats these outliers as a separate cluster.
Consequently, the latents generated by D$^4$M become concentrated around these outliers, leading to an \textbf{inaccurate representation of the VAE distribution}.
Besides, each individual \textcolor{orange}{\ding{56}} aims to map a single cluster, \textbf{rather than considering the entire subset of representative latents as a whole.}
Moreover, \red{\ding{72}} demonstrates that after adding noise and removing noise, \textbf{the latents undergo a shift from \textcolor{orange}{\ding{56}}}.
Therefore, these results align well with our previous discussion.

\subsection{More Analysis about Group Sampling}
\noindent \textbf{Computational efficiency}.~We report the runtime of the group sampling process across different $m$ and different GPUs in the table, demonstrating its simplicity and efficiency. As discussed in Lines 302-306, left column, this is because multiple subsets can be sampled in parallel on the GPU instead of sequential sampling, and the operations such as random sampling and mean/variance computations are very lightweight and efficient in current computation frameworks, making the entire process highly efficient.

\begin{table}[H]
\centering
\resizebox{0.8\linewidth}{!}{
\begin{tabular}{c|ccccc}
\toprule
IPC & 1 & 5 & 10 & 20 & 50 \\
\midrule
Time (s) & 0.3901$\pm$0.0057 & 1.0288$\pm$0.0092 & 1.8352$\pm$0.0012 & 3.506$\pm$0.0046 & 8.5252$\pm$0.0055 \\
\bottomrule
\end{tabular}
}
\caption{\textbf{Runtime on A100 40G with $m = 1e6$.}}
\label{tab:A1001e6}
\end{table}

\begin{table}[H]
\centering
\resizebox{0.8\linewidth}{!}{
\begin{tabular}{c|ccccc}
\toprule
IPC & 1 & 5 & 10 & 20 & 50 \\
\midrule
Time (s)& 0.1135$\pm$0.0016 & 0.1442$\pm$0.0013 & 0.2201$\pm$0.0003 & 0.3806$\pm$0.0046 & 0.9013$\pm$0.0060 \\
\bottomrule
\end{tabular}
}
\caption{\textbf{Runtime on A100 40G  with $m=1e5$.}}
\label{tab:A1001e5}
\end{table}

\begin{table}[H]
\centering
\resizebox{0.8\linewidth}{!}{
\begin{tabular}{c|ccccc}
\toprule
IPC & 1 & 5 & 10 & 20 & 50 \\
\midrule
Time (s) & 0.0902$\pm$0.0088 & 0.1670$\pm$0.0142 & 0.2783$\pm$0.0018 & 0.5227$\pm$0.0073 & 1.2873$\pm$0.0023 \\
\bottomrule
\end{tabular}
}
\caption{\textbf{Runtime on A6000 with $m=1e5$.}}
\label{tab:A60001e5}
\end{table}

\noindent \textbf{Increasing/decreasing Variance}.~As discussed in \cref{subsec:gaussian sampling}, we generate synthetic samples following the real data distribution, to ensure the training performance on synthetic samples. 
However, fow low IPCs, it is not feasible to fully match the original distribution.  An intuitive approach is to increasing/decreasing the variance yields a better target distribution.
We perform the experiment to verify the influence of variance. As shown in \cref{tab:variance}, we adjust the variance of the distribution by $±50\%$ in sampling. The results show that neither increasing nor decreasing the variance leads to higher accuracy. 
Therefore, changing the variance may lead to a distribution deviation with samples not similar to real ones and degraded training performance.

\begin{table}[H]
\centering
\begin{tabular}{cccccccc}
\toprule
$-50\%$ & $-30\%$ & $-10\%$ & $0$ & $+10\%$ & $+30\%$ & $+50\%$ \\
\midrule
39.8$\pm$0.5 & 42.2$\pm$0.4 & 43.1$\pm$0.4 & 44.1$\pm$0.3 & 41.4$\pm$0.4 & 37.4$\pm$0.5 & 34.0$\pm$0.6 \\
\bottomrule
\end{tabular}
\caption{\textbf{Impact of increasing/decreasing variance in group sampling.}}
\label{tab:variance}
\end{table}




\subsection{Visualization of Storage requirements}
\begin{figure}[H]
    \centering   \includegraphics[width=0.7\linewidth]{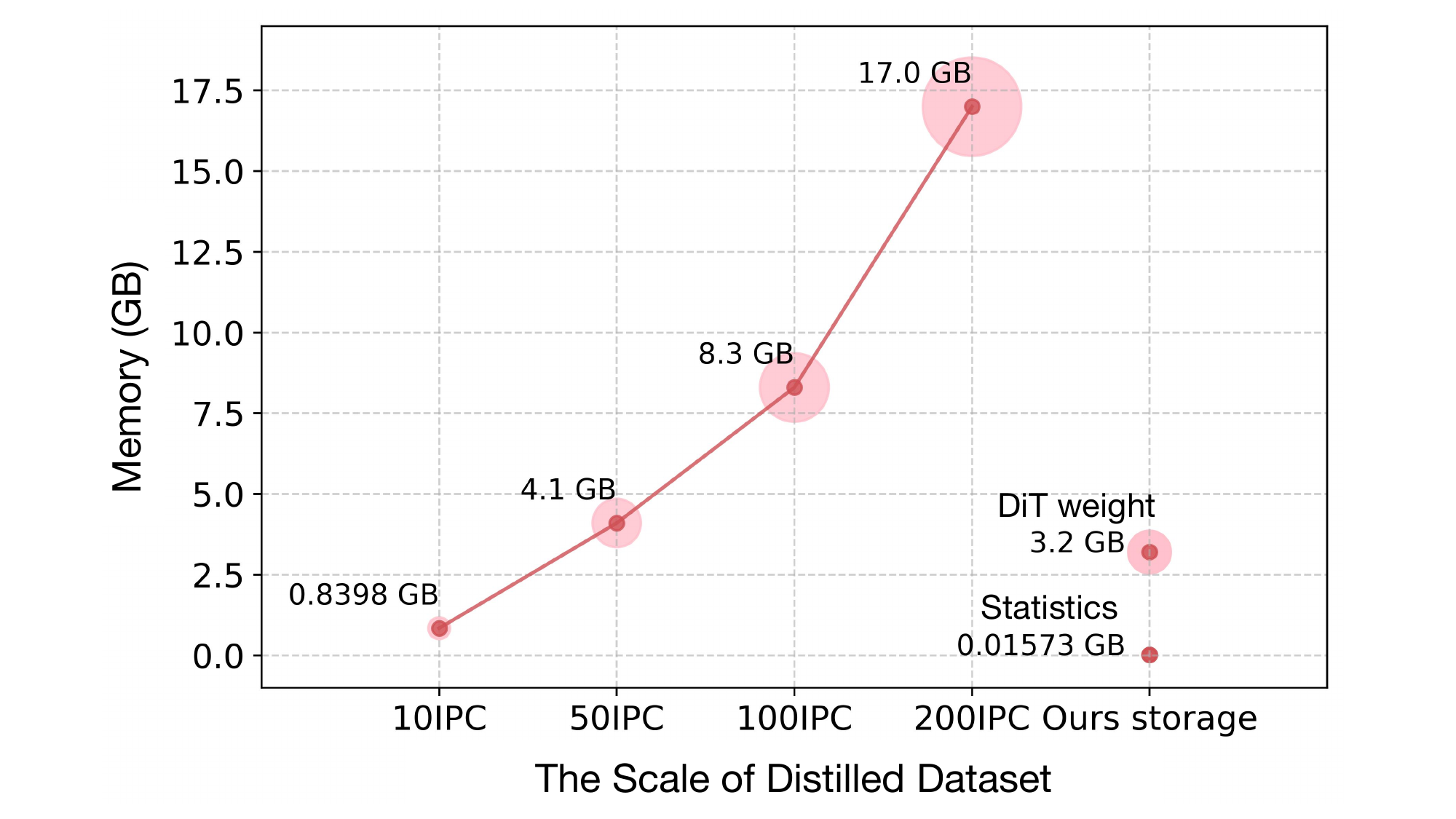}
    \caption{\textbf{The storage requirements of different settings on ImageNet-1K.}}
    \label{fig:storage}
\end{figure}

We present the storage requirements of different settings as discussed in \cref{subsec:Storage} here.
It can be seen in \cref{fig:storage}, the statistical parameters are quite small, requiring only about $0.016$ GB, which includes $320$ MB from the VAE weights.
The storage space needed for the distilled dataset scales linearly with the increase in IPCs.
When the IPC is large, the storage demand becomes significantly high.
However, the combined storage requirement for the DiT weights and statistical parameters remains smaller than that of a distilled dataset with $50$ IPC, yet it encapsulates all the essential information from the distillation process, regardless of IPC size.
Thus, this approach offers a highly efficient way to reduce storage consumption, especially for large IPCs.
Moreover, it can even encapsulate more images than the original dataset for improved training performance.

\subsection{Evolution of the Latent Space Across Timesteps}
\begin{figure}[H]
    \centering   \includegraphics[width=\linewidth]{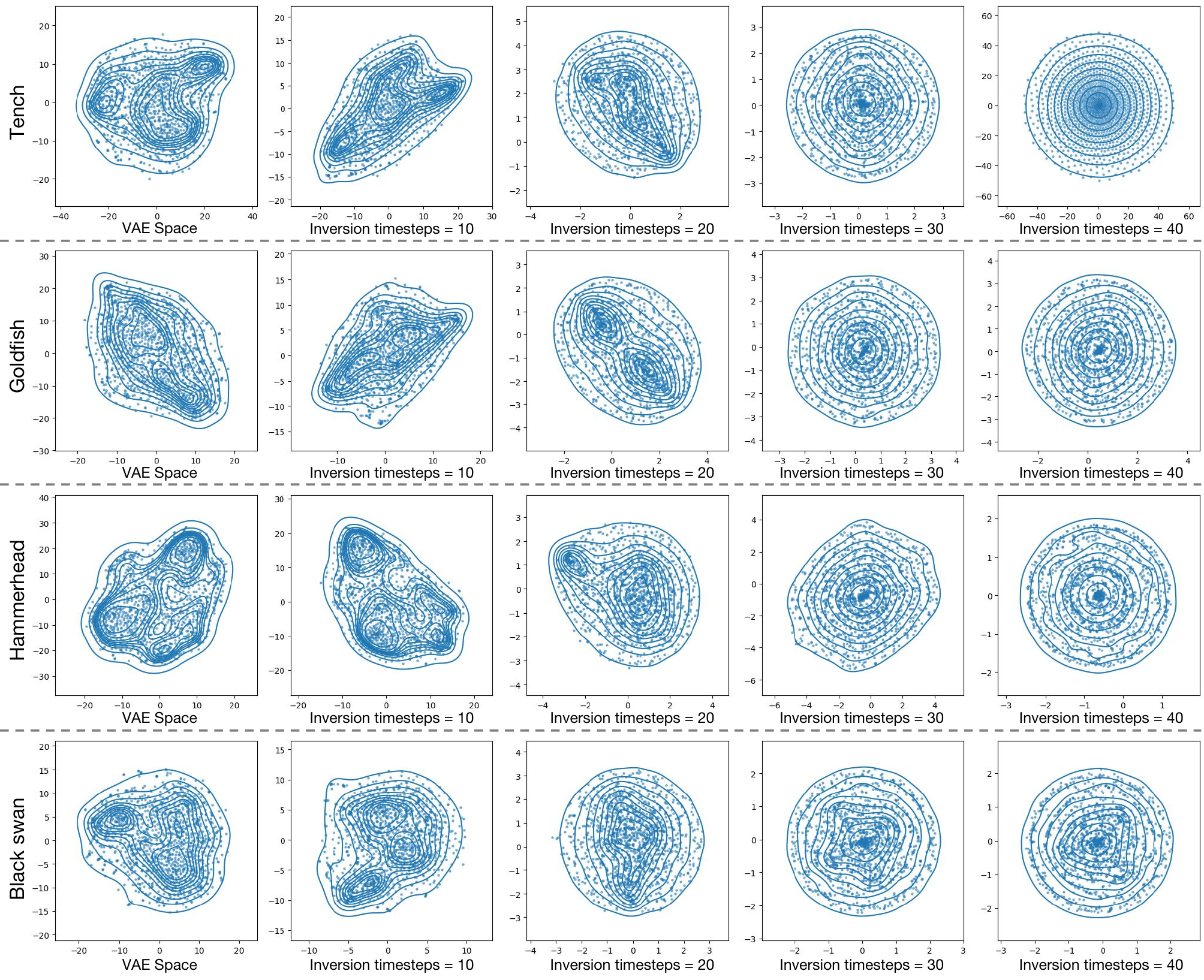}
    \caption{\textbf{T-SNE visualization of latent space at different inversion timesteps.}}
    \label{fig:5steps}
\end{figure}

We analyze the impact of different inversion timesteps on accuracy in \cref{subsec:different_steps}.
Here, we provide the visualization of the latent space transformation for different classes as inversion timestep increase \cref{fig:5steps}.
As demonstrated in \cref{sec:Theoretical}.2, the latent space progressively transforms into a high-normality Gaussian space with the increasing timesteps.

\subsection{More Comparison of Other SOTA Methods under Different Setting}

Some SOTA methods adopt different validation settings. Specifically, they train their validation models for $1000$ epochs on CIFAR-10 and CIFAR-100, whereas our main comparison uses $400$ training epochs.
For completeness, we provide a comparison under their settings on both CIFAR and ImageNet-1K, as shown in \cref{tab:more_com_cifar,tab:more_com_img}.
Our method consistently outperforms these approaches under their setup.

         


\begin{table}[H]
\centering
\begin{tabular}{c|c|c|cccc}
\toprule
Model & Dataset & IPC 
& G-VBSM & ESFR & HMDC & Ours \\
\midrule

\multirow{3}{*}{ResNet-18} 
  & CIFAR10 & 10 & 53.5$\pm$0.6 & 56.9$\pm$0.5 & 69.8$\pm$0.4 & \textbf{69.8$\pm$0.5} \\
  &                       CIFAR10   & 50 & 59.2$\pm$0.4 & 68.3$\pm$0.3 & 75.8$\pm$0.6 & \textbf{85.2$\pm$0.4} \\
  & CIFAR100                & 50 & 65.0$\pm$0.5 & -           & -           & \textbf{67.3$\pm$0.4} \\
\midrule

\multirow{3}{*}{ConvNetW128} 
  & CIFAR10 & 10 & 46.5$\pm$0.7 & -           & -           & \textbf{55.2$\pm$0.5} \\
  &                       CIFAR10  & 50 & 54.3$\pm$0.3 & -           & -           & \textbf{66.8$\pm$0.4} \\
  & CIFAR100                & 50 & 45.7$\pm$0.4 & -           & -           & \textbf{52.1$\pm$0.5} \\
\midrule

\multirow{3}{*}{ConvNet} 
  & CIFAR10 & 10 & -           & \textbf{59.9$\pm$0.2} & 47.5$\pm$0.7 & 57.3$\pm$0.4 \\
  &                       CIFAR10  & 50 & -           & 69.0$\pm$0.2 & 52.4$\pm$0.1 & \textbf{69.3$\pm$0.4} \\
  & CIFAR100                & 50 & -           & 51.3$\pm$0.4 & -           & \textbf{54.6$\pm$0.4} \\
\bottomrule

\end{tabular}
\caption{\textbf{Comparison with more SOTA methods on CIFAR-10 and CIFAR-100 (1000-epoch validation).}}
\label{tab:more_com_cifar}
\end{table}

\begin{table}[H]
\centering
\begin{tabular}{c|c|cccc}
\toprule
Model & IPC & G-VBSM & Teddy(post) & Teddy(prior) & Ours \\
\midrule
\multirow{3}{*}{ResNet-18} 
& 10  & 31.4$\pm$0.5 & 32.7$\pm$0.2 & 34.1$\pm$0.1 & \textbf{44.3}$\pm$\textbf{0.3} \\
& 50  & 51.8$\pm$0.4 & 52.5$\pm$0.1 & 52.5$\pm$0.1 & \textbf{59.4}$\pm$\textbf{0.1} \\
& 100 & 55.7$\pm$0.4 & 56.2$\pm$0.2 & 56.5$\pm$0.1 & \textbf{62.5}$\pm$\textbf{0.0} \\
\midrule
\multirow{3}{*}{ResNet-101} 
& 10  & 38.2$\pm$0.4 & 40.0$\pm$0.1 & 40.3$\pm$0.1 & \textbf{52.1}$\pm$\textbf{0.4} \\
& 50  & 61.0$\pm$0.4 & -- & -- & \textbf{66.1}$\pm$\textbf{0.1} \\
& 100 & 63.7$\pm$0.2 & -- & -- & \textbf{68.1}$\pm$\textbf{0.0} \\
\bottomrule
\end{tabular}
\caption{\textbf{Comparison with more SOTA methods on ImageNet-1k.}}
\label{tab:more_com_img}
\end{table}

\subsection{Cross-architecture Generalization}
We broaden our experiments in \cref{subsec:cross-architecture} by integrating diverse neural network architectures, including EfficientNet-B0 \cite{tan2019efficientnet}, ShuffleNet-V2 \cite{ma2018shufflenet}, DeiT-Tiny \cite{touvron2021training}.
It can be seen in \cref{tab:results}, \methodAbbr exhibits exceptional performance across substantial different architectures, incurring only a one-time generation cost.

\clearpage
\begin{table*}[t]
\centering
\resizebox{\textwidth}{!}{
\begin{tabular}{lc|cccccc}
\toprule
\multicolumn{2}{c|}{\textbf{Student\textbackslash Teacher}} & ResNet-18 & EfficientNet-B0 & MobileNet-V2 & ShuffleNet-V2 & VGG-11 & DeiT-tiny \\
\midrule
\multirow{2}{*}{ResNet-18}
 & RDED & $42.3 \pm 0.6$ & $\textbf{31.0} \pm \textbf{0.1}$ & $40.4 \pm 0.1$ & $43.3\pm0.3 $ & $36.6 \pm 0.1$ & $25.7\pm0.7$\\
 & Ours & $\textbf{44.2} \pm \textbf{0.3}$ & $30.6 \pm 0.2$ & $\textbf{42.3} \pm \textbf{0.7}$ & $\textbf{44.7} \pm \textbf{0.1}$ & $\textbf{38.3} \pm \textbf{0.2}$ & $\textbf{28.0} \pm \textbf{0.4}$\\
\midrule
\multirow{2}{*}{EfficientNet-B0}
& RDED & $42.8 \pm 0.5$ & $33.3 \pm 0.9$ & $43.6 \pm 0.2$ & $50.5\pm0.6$ & $35.8 \pm 0.5$ & $32.9\pm0.4$\\
& Ours & $\textbf{50.3} \pm \textbf{0.2}$ & $\textbf{42.0} \pm \textbf{0.6}$ & $\textbf{54.0} \pm \textbf{0.1}$ & $\textbf{56.3} \pm \textbf{0.1}$ & $\textbf{43.3} \pm \textbf{0.1}$ & $\textbf{36.8} \pm \textbf{0.4}$ \\
\midrule
\multirow{2}{*}{MobileNet-V2}
& RDED & $34.4 \pm 0.2$ & $24.1 \pm 0.8$ & $33.8 \pm 0.6$ & $48.2\pm0.5$ & $28.7 \pm 0.2$ & $24.9\pm0.6$\\
 & Ours & $\textbf{43.4} \pm \textbf{0.3}$ & $\textbf{31.8} \pm \textbf{0.8}$ & $\textbf{46.4} \pm \textbf{0.2}$ & $\textbf{50.0} \pm \textbf{0.2}$ & $\textbf{37.8} \pm \textbf{0.4}$ & $\textbf{26.5} \pm \textbf{0.8}$ \\
\midrule
\multirow{2}{*}{ShuffleNet-V2}
& RDED & $37.0\pm0.1$ & $23.7\pm1.0$ & $35.6\pm0.2$ & $40.5\pm0.0$ & $29.4\pm0.3$ & $21.8\pm0.6$\\
 & Ours & $\textbf{37.2} \pm \textbf{0.0}$ & $ \textbf{25.8} \pm \textbf{0.3}$ & $\textbf{38.5} \pm \textbf{0.4}$ & $\textbf{44.1} \pm \textbf{0.3}$ & $\textbf{32.6} \pm \textbf{0.6}$ & $\textbf{23.1} \pm \textbf{0.3}$ \\
\midrule
\multirow{2}{*}{VGG-11}
& RDED & $22.7 \pm 0.1$ & $16.5\pm 0.8$ & $21.6 \pm0.2$ & $25.7\pm0.4$ & $23.5 \pm 0.3$ & $\textbf{17.6}\pm\textbf{0.4}$\\
& Ours & $\textbf{25.7} \pm \textbf{0.4}$ & $\textbf{20.2} \pm \textbf{0.2}$ & $\textbf{24.8} \pm \textbf{0.4}$ & $\textbf{29.1} \pm \textbf{0.4}$ & $\textbf{28.1} \pm \textbf{0.1}$ & $15.1 \pm 0.2$ \\
\midrule
\multirow{2}{*}{DeiT-tiny}
& RDED & $13.2\pm0.3$ & $12.6\pm0.5$ & $13.6\pm0.4$ & $16.1\pm0.3$ & $11.4\pm0.1$ &
$\textbf{15.4}\pm\textbf{0.3}$\\
& Ours & $\textbf{17.1} \pm \textbf{0.3}$ & $ \textbf{14.8} \pm \textbf{0.2}$ & $\textbf{17.9} \pm \textbf{0.7}$ & $\textbf{21.4} \pm \textbf{0.8}$ & $\textbf{14.6} \pm \textbf{0.5}$ & $15.0 \pm 0.3$ \\
\bottomrule
\end{tabular}
}
\caption{\textbf{Comparison of Top-1 accuracy for cross-architecture generalization on ImageNet-1K, $IPC=10$.}
}
\label{tab:results}
\end{table*}

\subsection{More Ablation Studies on the Hyper-parameters}

\noindent \textbf{Choice of $m$ across Different Datasets.} As shown in \cref{tab:m_sweep1,tab:m_sweep2}, we present the performance of varying $m$ on datasets of varying scales.
As expected, increasing $m$ improves accuracy, as a larger candidate pool offers greater diversity and a higher chance of including representative sets. However, when $m$ becomes sufficiently large, the performance gains plateaus—indicating that the marginal benefit of adding more candidates diminishes, as the top-performing candidates become increasingly similar. We choose $m$ at the saturation point, typically between $1e5$ and $1e7$, which can work well across different datasets.

\begin{table}[H]
\centering
\begin{tabular}{c|ccccccc}
\toprule

$m$   & 1   & $1\text{e}3$ & $1\text{e}4$ & $1\text{e}5$ & $5\text{e}5$ & $1\text{e}6$ & $5\text{e}6$ \\
\midrule
Accuracy (\%)   & 40.2$\pm$0.4 & 40.9$\pm$0.4 & 41.7$\pm$0.5 & 42.4$\pm$0.4 & 43.8$\pm$0.3 & 44.2$\pm$0.2 & 44.1$\pm$0.3 \\

\bottomrule
\end{tabular}
\caption{\textbf{The results across different $m$ under $IPC=10$ on Tiny-ImageNet.}}
\label{tab:m_sweep1}
\end{table}

\begin{table}[H]
\centering
\begin{tabular}{c|ccccccc}
\toprule
$m$   & 1   & $1\text{e}3$ & $1\text{e}5$ & $1\text{e}6$ & $5\text{e}6$ & $1\text{e}7$ & $5\text{e}7$ \\
\midrule
Acc (\%)   & 38.9$\pm$0.3 & 39.6$\pm$0.2 & 40.4$\pm$0.3 & 41.0$\pm$0.1 & 41.8$\pm$0.1 & 41.9$\pm$0.2 & 42.1$\pm$0.2 \\
\bottomrule
\end{tabular}
\caption{\textbf{The results across different $m$ under $IPC=10$ on CIFAR-10.}}
\label{tab:m_sweep2}
\end{table}

\noindent \textbf{Different Combination of $\lambda_{\mu}$, $\lambda_{\sigma}$, $\lambda_{\gamma_1}$.} For $\lambda_\mu$, $\lambda_\delta$ and $\lambda_\gamma$, we set them to make the corresponding metrics on the same scale. We provide results for different $\lambda$ in \cref{tab:setting_ratios}, which shows that the performance on $L_{T,C}$ is relatively robust to $\lambda$.

\begin{table}[H]
\centering
\begin{tabular}{ccccc}
\toprule
Setting & 1:1:0.5 & 1:1:1 & 1:1:2 & 1:0.5:0.5 \\
\midrule
Acc (\%) & 44.2$\pm$0.1 & 44.0$\pm$0.4 & 43.6$\pm$0.3 & 43.4$\pm$0.2 \\
\bottomrule
\end{tabular}
\label{tab:setting_ratios}
\caption{\textbf{The results of different combination of $\lambda_{\mu}$, $\lambda_{\sigma}$, $\lambda_{\gamma_1}$.}}
\end{table}

\section{Image Visualization}   \label{sec:ad_Visualization}
\begin{figure}[t]
    \centering
    \includegraphics[width=\linewidth]{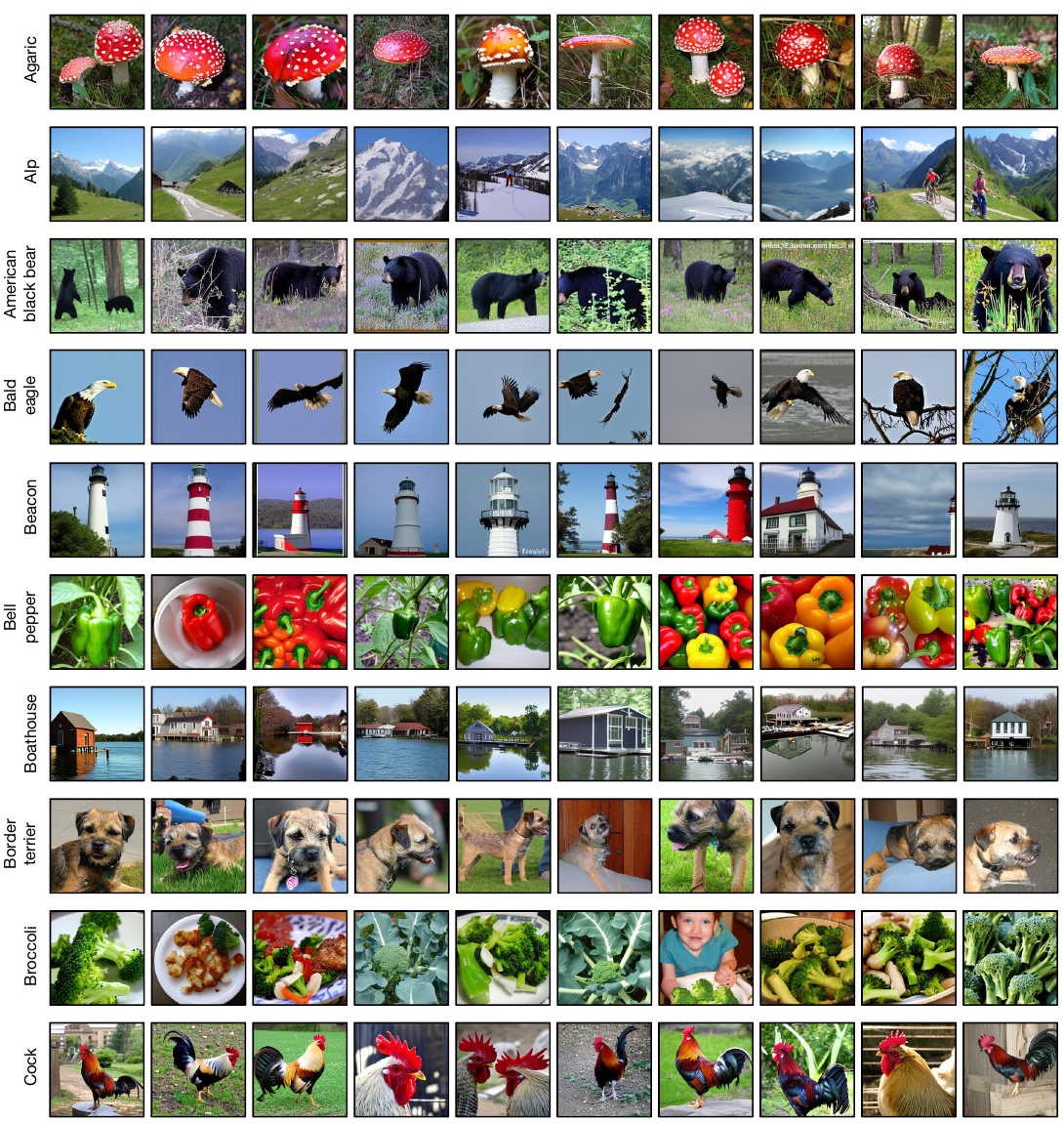}
    \caption{\textbf{Visualization of the distilled images for different classes on ImageNet-1K, $IPC=10$, and the resolution is $224 \times 224$.}}
    \label{fig:appendix1}
\end{figure}
\begin{figure}[t]
    \centering
    \includegraphics[width=\linewidth]{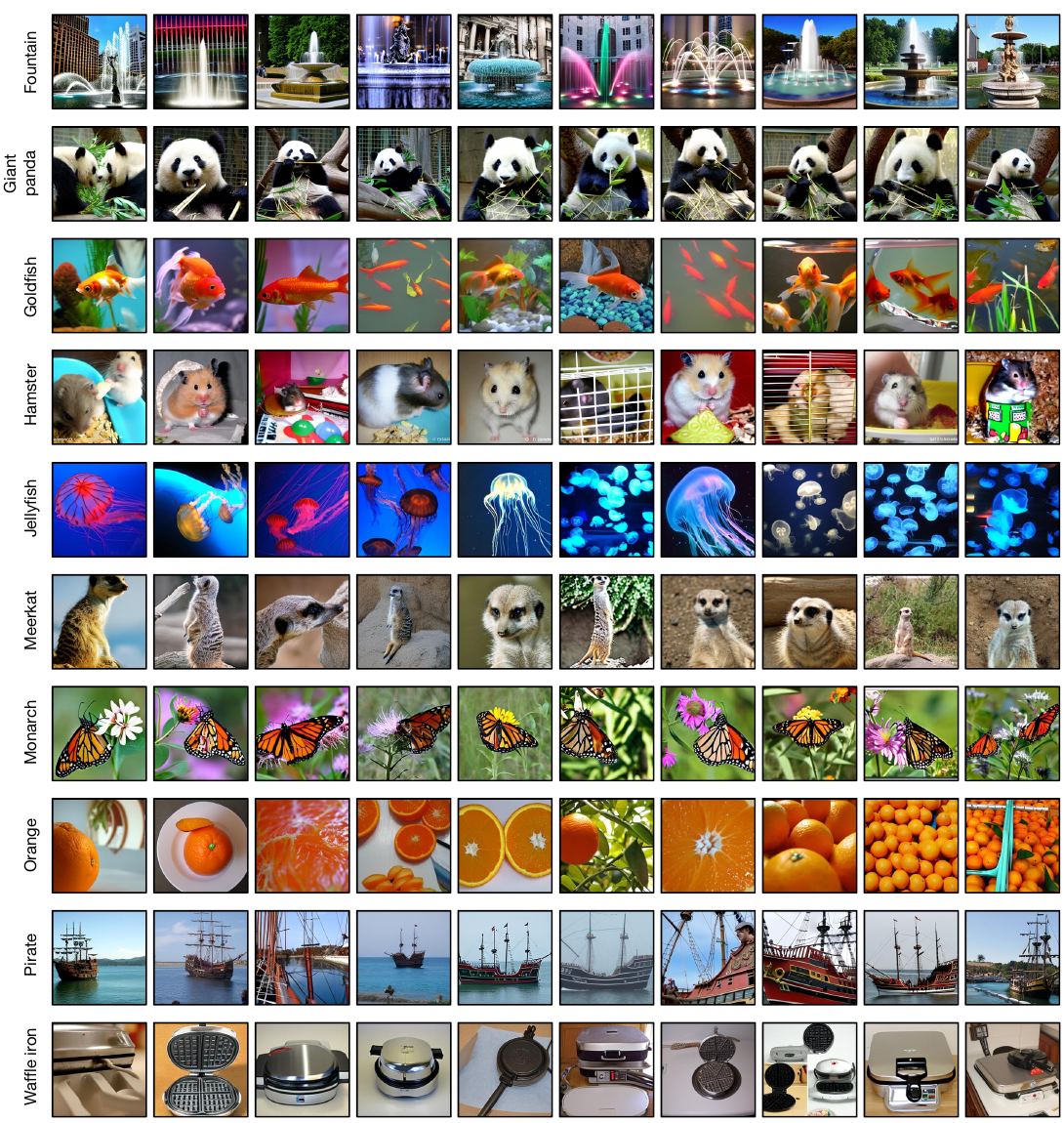}
    \caption{\textbf{Visualization of the distilled images for different classes on ImageNet-1K, $IPC=10$, and the resolution is $224 \times 224$.}}
    \label{fig:appendix2}
\end{figure}
We present more visualization results of the distilled images in this section.
As shown in \cref{fig:appendix1} and \cref{fig:appendix2}, \methodAbbr generates diversity, high-quality images for each class, effectively representing the full dataset.

\end{document}